\documentclass[dvipsnames, 11pt]{article}

\usepackage[preprint]{acl}

\usepackage{times}
\usepackage{latexsym}

\usepackage[T1]{fontenc}

\usepackage[utf8]{inputenc}

\usepackage{microtype}
\usepackage{url}
\usepackage{hyperref}
\usepackage{array}
\usepackage{amsmath}
\usepackage{amsthm}
\usepackage[inline]{enumitem}
\usepackage{multirow}
\usepackage{floatrow}
\usepackage{subcaption}
\usepackage{color, colortbl}
\usepackage{xcolor}
\usepackage[most]{tcolorbox}
\usepackage{inconsolata}
\usepackage{xspace}
\usepackage{floatrow}
\usepackage{algorithm}
\usepackage{algpseudocode}
\usepackage{soul}
\usepackage{fancyvrb}
\usepackage{listings}
\usepackage{booktabs}
\usepackage{graphicx}
\usepackage{amsfonts}
\hypersetup{
    colorlinks = true,
    linkcolor = Bittersweet,
    citecolor = Fuchsia
}
\newfloatcommand{capbtabbox}{table}[][\FBwidth]
\newfloat{lstfloat}{htbp}{lop}
\floatname{lstfloat}{Prompt}

\newfloat{lstdoc}{htbp}{lop}
\floatname{lstdoc}{Guideline}

\makeatletter
\def\@fnsymbol#1{\ensuremath{\ifcase#1\or \dagger\or \ddagger\or
   \mathsection\or \mathparagraph\or \|\or **\or \dagger\dagger
   \or \ddagger\ddagger \else\@ctrerr\fi}}

\title{Prototypical Human-AI Collaboration Behaviors\\ from LLM-Assisted Writing in the Wild}

\author{Sheshera Mysore\textsuperscript{1}$^{\bigtriangleup}$
\quad
Debarati Das\textsuperscript{2}\thanks{~~Work done during an internship at Microsoft}\quad{\bf Hancheng Cao\textsuperscript{1}}\quad
{\bf Bahar Sarrafzadeh\textsuperscript{1}$^{\bigtriangleup}$}\\
\textsuperscript{1}Microsoft, \textsc{\small WA, USA}
\textsuperscript{2}University Of Minnesota, \textsc{\small MN, USA}\\
$^\bigtriangleup$Corresponding authors:\\ \texttt{smysore@iesl.cs.umass.edu}, \texttt{bahar.sarrafzadeh@microsoft.com}
  }

\begin{document}
\newcommand{\proposedacronym}{\textsc{Path}\xspace}
\newcommand{\bcpwr}{{BCP$_{\text{Wr}}$}\xspace}
\newcommand{\wildchatwr}{WC$_{\text{Wr}}$\xspace}
\newcommand{\bcpall}{{BCP$_{\text{All}}$}\xspace}
\newcommand{\wildchatall}{\textsc{WC}$_{\text{All}}$\xspace}
\newcommand{\followupclf}{$f_{\text{FollowU}}$\xspace}
\newcommand{\fineintclf}{$f_{\text{WritingI}}$\xspace}
\newcommand{\coarsetaskclf}{$f_{\text{CoarseT}}$\xspace}
\newcommand{\orexample}[1]{\textcolor{MidnightBlue!90}{#1}}
\newcommand{\fupexample}[1]{\textcolor{Blue!60}{#1}}
\definecolor{greeni}{HTML}{9ec5aa}
\definecolor{neonbluei}{HTML}{49f0fa}
\definecolor{redi}{HTML}{ec6d57}
\definecolor{purplei}{HTML}{8400ed}
\newcommand{\greenintent}[1]{\colorbox{greeni!42}{#1}}
\newcommand{\neonblueintent}[1]{\colorbox{neonbluei!42}{#1}}
\newcommand{\redintent}[1]{\colorbox{redi!42}{#1}}
\newcommand{\purpleintent}[1]{\colorbox{purplei!42}{#1}}

\maketitle

\begin{abstract}
As large language models (LLMs) are used in complex writing workflows, users engage in multi-turn interactions to steer generations to better fit their needs. Rather than passively accepting output, users actively refine, explore, and co-construct text. We conduct a large-scale analysis of this collaborative behavior for users engaged in writing tasks in the wild with two popular AI assistants, Bing Copilot and WildChat. Our analysis goes beyond simple task classification or satisfaction estimation common in prior work and instead characterizes how users interact with LLMs through the course of a session. We identify prototypical behaviors in how users interact with LLMs in prompts following their original request. We refer to these as \ul{P}rototypical Human-\ul{A}I Collabora\ul{t}ion Be\ul{h}aviors (\proposedacronym{}s) and find that a small group of \proposedacronym{}s explain a majority of the variation seen in user-LLM interaction. These \proposedacronym{}s span users revising intents, exploring texts, posing questions, adjusting style or injecting new content. Next, we find statistically significant correlations between specific writing intents and \proposedacronym{}s, revealing how users' intents shape their collaboration behaviors. We conclude by discussing the implications of our findings on LLM alignment.
\end{abstract}

\section{Introduction}
\label{sec-intro}
\begin{figure}
  \includegraphics[width=1\textwidth]{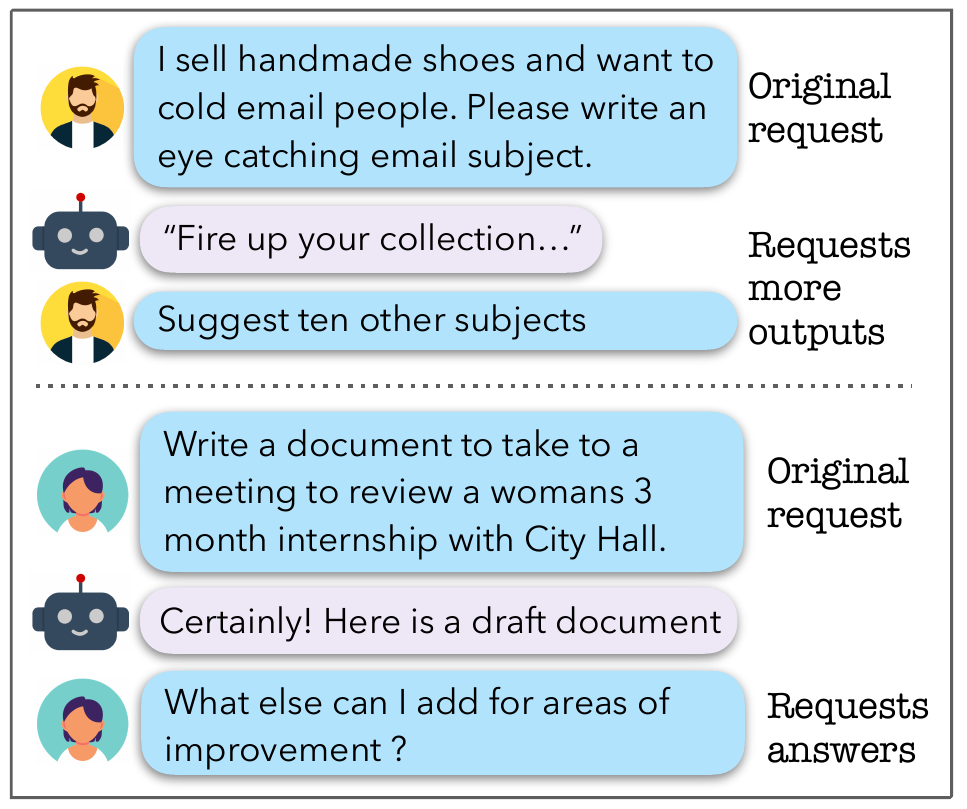}
  \caption{Users follow up their original requests to collaborate with LLMs in writing sessions. We identify prototypical human-AI collaboration behaviors (\proposedacronym{}s), and find statistically significant correlations between users' writing intents and \proposedacronym{}s.}
  \label{fig-teaser-examples}
\end{figure}
LLMs' generalization ability and natural language interfaces have made powerful AI models accessible to a range of users engaged in diverse tasks \citep{ouyang2022humanprefs}. The logged natural language interactions from LLM-powered AI assistants have emerged as a rich data source for understanding user-AI interaction \citep{zhu2025logchiea}. Leveraging this data, recent studies have explored the high-level tasks users engage in (e.g., search, coding, writing, etc.) \citep{tamkin2024clio}, their variation across occupations \cite{handa2025economic}, and measured user satisfaction based on interaction patterns \citep{lin2024interpretablesat}. However, little prior work have examined how users collaborate with LLMs in real-world LLM deployments. 

A notable characteristic of current AI assistants is their conversational nature, which enables users to engage in follow-up interactions after stating their original requests. These follow-ups allow users to articulate their needs better and obtain more helpful responses from the LLM (Figure \ref{fig-teaser-examples}). Analyzing these follow-up interactions promises to provide a rich characterization of human-AI collaboration and guide research on LLM alignment based on realistic in-the-wild interactions.

We contribute such an analysis by focusing on LLM-assisted writing, a important and increasingly prevalent use case for AI assistants \citep{tamkin2024clio, suri2024binganalysis}. Recent studies have found that LLM-assisted writing is now common in impactful domains such as press releases, job postings, and peer reviews, among others \citep{liang2025wideadopt, liang2024aiacademic}. Despite this, no prior research has systematically analyzed how users collaborate with AI assistants for writing tasks in the wild. We address this gap by formulating two key research questions: \textbf{(RQ1)} What high-level collaboration behaviors emerge from user interactions in AI-assisted writing? And \textbf{(RQ2)} How do these collaboration behaviors differ across writing intents?

To address these questions, we conduct a large-scale analysis of writing sessions from two AI assistants: Bing Copilot \citep{mehdi2023bingcopilot} and WildChat \citep{zhao2024wildchat}. Our datasets span 20.5M and 800k English user-LLM conversation sessions over seven and thirteen months of global Bing Copilot and WildChat usage, respectively. The two datasets enable us to identify shared collaboration behaviors across distinct AI assistants, and the public WildChat logs support reproducibility and future research. To answer RQ1, we use GPT-4o to classify users' follow-up utterances into high-level types and cluster them using Principal Component Analysis (PCA) \citep{bengio2013replearn}. Each cluster represents sessions with consistent collaboration behavior, which we term \ul{P}rototypical Human-\ul{A}I Collabora\ul{t}ion Be\ul{h}aviors (\proposedacronym{}). To address RQ2, we use GPT-4o to identify writing intents from the users original requests and conduct regression analysis to correlate them with \proposedacronym{}s. This lets us detect statistically significant relationships between writing intents and collaboration behaviors.

\noindent \textbf{Takeaways:} 
We identify seven \proposedacronym{}s that capture 80-85\% of variance across datasets, with shared behaviors like revising intents, exploring texts, asking questions, or modifying generations despite differences in deployments. Correlating writing intents and \proposedacronym{}s enables us to uncover intent-specific alignment needs. For instance, users sought to explore diverse generations in follow-ups in brainstorming eye-catching texts, indicating users need for LLMs aligned for brainstorming applications. Users generated long texts by staging generation and interactively providing feedback with different levels of specificity, indicating the need for session-level alignment from under-specific feedback. And in generating professional or technical texts, users followed up with questions aimed at learning about a domain's norms or to seek feedback, indicating the need to align LLMs for promoting learning in users. By analyzing collaborative writing behaviors in the wild, we offer insights to guide future research on LLM alignment.
\section{Related Work}
\label{sec-relatedwork}
The rise of interactive LLM systems has seen the emergence of user-LLM interaction log datasets \citep{kirk2024prism, zhao2024wildchat} and analysis, building on rich traditions of log analysis in HCI and Information Retrieval \citep{jansen2006ninesearchlogs, dumais2014loganalysis}. This work has developed an understanding of how users interact with intelligent systems in-the-wild aiming to inform future system and model development. Such work has analyzed user-LLM interactions in the context of information seeking \citep{trippas2024bardlogs}, theorem proving \citep{katherine2024mathinteractions}, image generation \citep{palmini2024civiversedatasetanalyzinguser, vodrahalli2024artwhisperer}, and writing assistance \citep{lee2022coauthor}. 
Most relevant is the prior work on log analysis for UI based writing assistance \citep{lee2022coauthor, sarrafzadeh2021stageaware}. Respectively, they analyze model outputs, collaboration patterns, and the ability of interactions to predict the different stages of writing in UI based applications. Our work differs in its examination of conversation logs to uncover collaboration behaviors. Our focus on user-LLM conversations ties to recent large scale analysis of such logs to infer high level tasks of conversations \citep{tamkin2024clio, suri2024binganalysis}.
We extend this beyond analyzing tasks alone by focusing on collaboration behaviors captured in users' follow-up utterances. Further, correlating collaboration behaviors with intents enables us to uncover meaningful implications for LLM alignment (\S\ref{sec-discussion}). Finally, \citet{katherine2024mathinteractions} present a notable exception in conducting a small-scale qualitative analysis of user-LLM collaboration behaviors in mathematical theorem proving. Similar to our findings (\S\ref{sec-results-moreoutputs},\ref{sec-results-followupq}), they find users to engage in exploration and question asking behaviors. Our work differs in its focus on writing, at scale analysis of in-the-wild interactions and, correlating writing intents with collaboration behaviors. We review related work on LLM-assisted writing with a human-centered focus and user satisfaction estimation from logs in Appendix \ref{sec-relatedwork-supp}, and further discuss relevant studies in the context of our results in Section \ref{sec-results}.

\section{Analysis Setup}
\label{sec-analysis-setup}
Our analysis is based on user-LLM conversational logs from Bing Copilot \citep[BCP]{mehdi2023bingcopilot} and WildChat-1M \citep[WC]{zhao2024wildchat}. The two systems vary in their base LLMs, interfaces, and user bases and allow our analysis to identify shared behaviors likely to hold beyond deployments. Further, the public WildChat-1M dataset enables reproducibility and future work based on our analysis. For our analysis, we focus on English sessions engaged in writing tasks, excluding sessions focused on tasks like search or software development. We conceptualize writing sessions broadly to be the ones where users generated inter-personal or public communicative texts, technical texts, creative texts, and those focused on summarization. We treat complete generations and rewrites of whole or parts of texts as writing. We operationalize our definition in an iteratively developed and manually validated GPT-4o based multi-label Task Classifier (\coarsetaskclf) and use it to identify sessions focused on writing. We refer to the writing log datasets as \bcpwr and \wildchatwr. Appendix \ref{sec-analysis-setup-supp} details both datasets, their filtering, and \coarsetaskclf.
 \begin{table}[t]
    \scalebox{0.8}{
    \begin{tabular}{rccl}
    \toprule
     & Users & Countries & Top 5 countries\\
    \midrule
    \bcpwr & 202k & $219$     &    US $30\%$, IN $15\%$, GB $6\%$\\
    & & & PH $6\%$, AU  $6\%$\\
    \wildchatwr & 22k &   $166$    & US $25\%$, GB $10\%$, RU $9\%$\\
    & & & IN $5\%$, PH $4\%$\\
    \bottomrule
    \end{tabular}
    }
    \caption{The number of users, countries, and countries where sessions originate (in \%) in \bcpwr and \wildchatwr. They have 250k and 68k sessions respectively.}
    \label{tab-exploratory-countries}
\end{table}

\textbf{Bing Copilot - \bcpwr} We construct \bcpwr from a daily random sample of sessions from Bing Copilot gathered from April-Oct 2024, resulting in 20.5M sessions. To enable the study of user-LLM collaboration, we ensure that sessions contain users' follow-up utterances and retain sessions with at least 2 user utterances and those in English. On the resulting 2.8M sessions, we run \coarsetaskclf to identify writing sessions and retain 250k sessions in \bcpwr. GPT-4 powered all interactions in BCP.

\textbf{WildChat - \wildchatwr} We follow a similar procedure to construct \wildchatwr  from the public WildChat-1M. \citet{zhao2024wildchat} gathered in a research study from April 2023 to May 2024. We retain sessions with at least 2 user utterances and English sessions. This results in 160k sessions of which 68k are identified as writing sessions by \coarsetaskclf. GPT-4 and GPT-3.5-Turbo powered the interactions in WC.

In Table \ref{tab-exploratory-countries}, we see that the resulting \bcpwr and \wildchatwr contain sessions from a large group of users from over 150 countries. While many sessions originate in English-speaking countries, we observe a long tail of countries. In Table \ref{tab-exploratory-session-stats}, we present session length characteristics, and find that sessions had 2 follow-ups on average and that half of all sessions contain only one writing intent.
\section{Log Analysis with \proposedacronym{}s}
\label{sec-method}
\begin{figure}
  \includegraphics[width=0.9\textwidth]{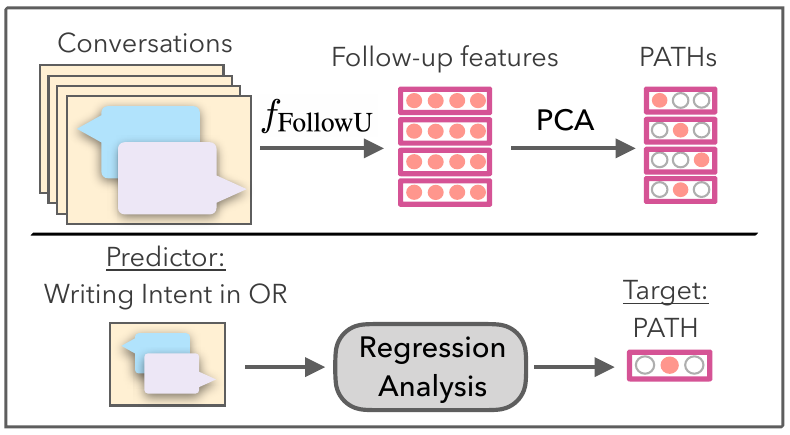}
  \caption{Our analysis methods aim to discover \proposedacronym{}s (above) and identify how \proposedacronym{}s differ across writing intents in original requests (ORs) (below).}
  \label{fig-analysis-overview}
\end{figure}
Our analysis centers around two research questions: (RQ1) What high-level collaboration behaviors emerge from user interactions in LLM-assisted writing? And (RQ2) How do these collaboration behaviors differ across writing intents? We overview our analysis method here (see Figure \ref{fig-analysis-overview}) and include detailed descriptions in Appendix \ref{sec-method-supp}.

\textbf{Identifying PATHs.} To answer RQ1 we classify user utterances into ``Original Requests'' (OR) and set of follow-up types (Table \ref{tab-follow-up-descr}) using a GPT-4o based follow-up classifier (\followupclf). ORs represent the primary writing goal of a user in a session, and follow-up types categorize users' follow-ups into higher-level behaviors (e.g., Figure \ref{fig-teaser-examples}). \followupclf was developed iteratively and manually validated to ensure its accuracy for subsequent analysis (see Appendix \ref{sec-method-identifypref-supp}). Then, we identify co-occurring patterns of follow-up types using Principal Component Analysis (PCA) \citep{bengio2013replearn}. We take each principal component to represent a \proposedacronym{}. 
\begin{figure*}
\begin{floatrow}
  \capbtabbox{%
  \scalebox{0.7}{
    \begin{tabular}{rl}
    \toprule
    User utterance type & Description\\
    \midrule
    \textsc{original request} & User makes a new request\\
    \textsc{restates request} & Reformulates their request\\
    \textsc{elaborates request} & Expands on their request\\
    \textsc{requests answers} & Question related to output\\
    \textsc{requests more outputs} & Asks for additional output\\
    \textsc{change style} & Changes style of output\\
    \textsc{adds content} & Adds content to output\\
    \textsc{removes content} & Remove content from output\\
    \textsc{courtesy response} & A courtesy or pleasantry\\
    \textsc{responds positively} & Explicitly pleased with output\\
    \textsc{responds negatively} & Explicitly unhappy with output\\
    \textsc{undefined response} & No defined label applies\\
    \bottomrule
    \end{tabular}}}
    {\caption{User utterances are classified into \textsc{original requests} and high-level follow-up types.} 
\label{tab-follow-up-descr}}
 \capbtabbox{%
    \scalebox{0.7}{
        \begin{tabular}{ll}
        \toprule
        \multicolumn{2}{c}{Writing Intent Types}\\
        \midrule
        \textsc{improve text} & \textsc{generate professional doc}\\
        \textsc{generate message} & \textsc{generate catchy text}\\
        \textsc{generate bio} & \textsc{generate story}\\
        \textsc{generate summary} & \textsc{generate technical text}\\
        \textsc{generate script} & \textsc{generate character}\\
        \textsc{generate essay} & \textsc{generate poem}\\
        \textsc{get references} & \textsc{generate song}\\
        \textsc{generate online post} & \textsc{generate joke}\\
        \textsc{question about writing} & \textsc{undefined request}\\
        \bottomrule
        \end{tabular}
        }
    }{\caption{Users \textsc{original requests} are classified into the above writing intents.} 
    \label{tab-fgtype-descr}
    }
\end{floatrow}
\end{figure*}

Specifically, we represent a dataset's sessions $\mathcal{S}$ with a ``\texttt{tf-idf}'' representation of its follow-up types ($\mathbf{F}$) and run PCA on $\mathbf{F}$. This transforms it as $\mathbf{P} = \mathbf{F}\mathbf{W}$ where each dimension of $\mathbf{P}$ represents a mutually co-occurring set of follow-up types and segments $\mathcal{S}$ into subsets of sessions with a consistent \proposedacronym{}. We retain the first $l$ dimensions of $\mathbf{P}$ that explain 80-85\% variance in $\mathbf{F}$, treating the rest as noise. Our use of PCA follows its standard use for exploratory data analysis \citep{eagle2009eigenbehaviors, reades2009eigenplaces}, its linear nature enables us to easily visualize how follow-up types combine to form \proposedacronym{}s (through $\mathbf{W}$), and its closed form solution ensures the consistency of results across re-runs \citep{greene2014many}. 

\textbf{Correlating Intents and \proposedacronym{}s.} To answer RQ2, we classify Original Requests into a finer-grained set of writing intents (Table \ref{tab-fgtype-descr}) with a GPT-4o based multi-label intent classifier (\fineintclf). \fineintclf was also developed iteratively and manually validated (see Appendix \ref{sec-method-correlatepref-supp}). Then we run a logistic regression correlating intents (predictors) from \fineintclf with \proposedacronym{}s (targets). Analyzing the learned coefficients of the regression models allows us to identify statistically significant correlations between writing intents and PATHs in a principled manner. Logistic regressions also enable easy interpretation and follow on a large body of prior work \citep{gujarati2021essentials}. Finally, to gain a deeper understanding of user behaviors in a correlated intent-\proposedacronym{} pair, two authors conducted a qualitative analysis of the pairs, which showed statistically significant correlations and were repeated in \bcpwr and \wildchatwr. Author-driven manual analysis was conducted to overcome the lack of access to Bing Copilot or WildChat users, a fundamental challenge in all log-based studies \citep{dumais2014loganalysis}. In \S\ref{sec-results} we include example conversations examined by both authors to illustrate our findings.
\section{Results -- Exploring \proposedacronym{}s}
\label{sec-expanalysis}
\begin{figure*}[t]
\captionsetup[subfigure]{justification=centering}
\centering
\begin{subfigure}[t]{.18\textwidth}
    \centering 
    {\includegraphics[width=1.14\textwidth]{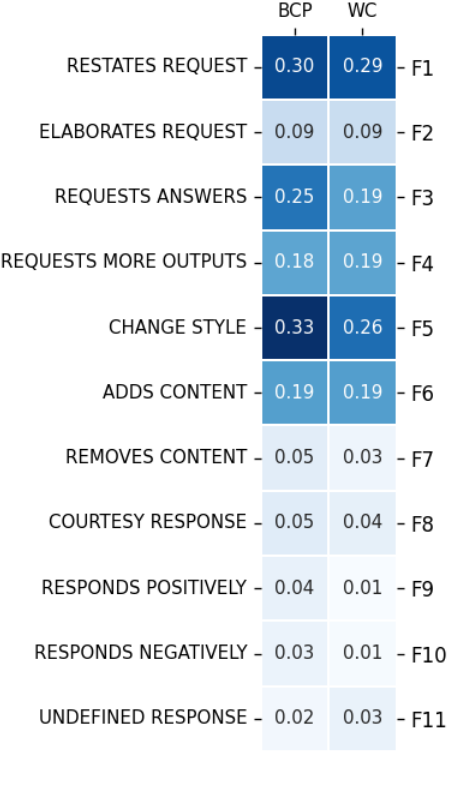}}
    \caption{Frequencies.}
    \label{fig-bothfupfreq}
\end{subfigure}\quad
\begin{subfigure}[t]{.77\textwidth}
    \centering
    {\includegraphics[width=\textwidth]{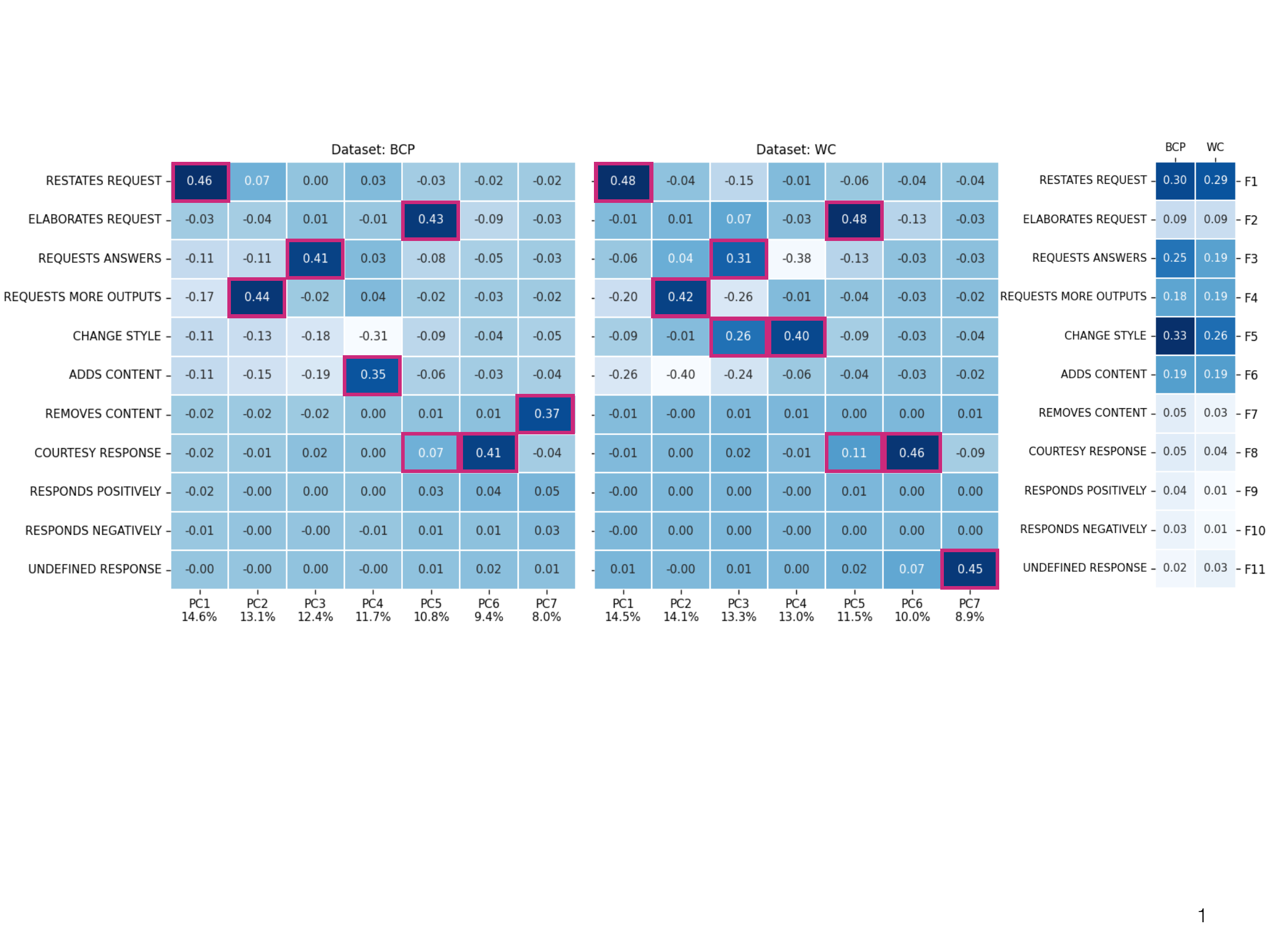}}
    \caption{Correlations between follow-up types and PCs in \bcpwr and \wildchatwr.}
    \label{fig-bothloadings}
\end{subfigure}
\caption{(a) The fraction of sessions which contain a follow-up type. (b) The correlation between follow-up types and principal components (PC) inferred by PCA. Each PC represents a \proposedacronym{}. Large positive values (pink boxes) indicate stronger correlations. The percentages (bottom) depict the variance explained by each \proposedacronym{}.} 
\label{fig-followupexp}
\end{figure*}

We start with (RQ1): What high-level collaboration behaviors emerge from user interactions in LLM-assisted writing? We do this by visualizing the frequency of follow-up types identified by \followupclf (Figure \ref{fig-bothfupfreq}) and the correlations between the follow-up types and \proposedacronym{}s identified by PCA (Figure \ref{fig-bothloadings}). We discuss specific \proposedacronym{}s alongside writing intents and examples in \S\ref{sec-results}.

\textbf{Follow-up Trends.} 
In Figure \ref{fig-bothfupfreq}, we see that the most frequent follow-up types across \bcpwr and \wildchatwr are similar (Table \ref{tab-follow-up-examples} contains examples). Across datasets, users frequently ($18$-$30$\% sessions) follow up by revising or elaborating on their Original Request (F1, F2 in Fig.\ \ref{fig-bothfupfreq}), ask questions about the generation (F3), explore additional outputs (F4), or modify the generations (F5, F6). Follow-ups with explicit positive/negative feedback or courtesy responses indicating satisfaction (F8-F10) are rare and occur only in $1$-$5$\% of the sessions.

\textbf{High-level trends in \proposedacronym{}s.} 
In Figure \ref{fig-bothloadings} we visualize how follow-up types form \proposedacronym{}s ($\mathbf{W}$ from PCA) and the variance explained by each \proposedacronym{}/PC. Seven \proposedacronym{}s explain 80-85\% of the variance, showing that a small set of collaboration behaviors explains the bulk of variance in follow-up behaviors in \bcpwr and \wildchatwr. Further, each \proposedacronym{} accounts for a similar and small percentage of variance (8-14\%), with most \proposedacronym{}s corresponding to a single follow-up type. This suggests that each follow-up type captures a distinct form of collaboration, with multiple co-occurring follow-up types being less frequent. When follow-up types do co-occur, this is more common in rare follow-up types (e.g., \textsc{elaborates request} and \textsc{courtesy response} co-occur in \proposedacronym{}5).

\textbf{\proposedacronym{}s in \bcpwr vs \wildchatwr.} 
In Figure \ref{fig-bothloadings}, we also see that the discovered \proposedacronym{}s share significant similarities despite differences in Bing Copilot and WildChat deployments (e.g.\ base LLMs, system prompts, interfaces, and user bases). Specifically, we find \textsc{restates request} (F1) accounts for similar amounts ($14.5$-$14.6\%$) and the maximum variance in both datasets. Revising requests is similar to query reformulations in search, such as Google or Bing Search. This is a dominant mode of interaction that is familiar to users \citep[Sec 7.4]{alaofi2022wherequeries}, and they continue to engage in it. 
Further, exploring additional outputs (\proposedacronym{}2) explains the 2nd largest amount of variance in both datasets. Similarly, \proposedacronym{}5 and \proposedacronym{}6 are also shared across both datasets, though these constitute less frequent follow-up types. These trends suggest that users collaborate with LLMs in very similar ways across both systems we examined. However, there are also some notable differences. \proposedacronym{}3 while correlating with \textsc{requests answers} in both datasets, also correlated with \textsc{change style} in \wildchatwr. Similarly, \proposedacronym{}4 correlated with \textsc{adds content} and \textsc{change style} respectively. Despite the differences in \proposedacronym{}3 and 4, note that they aim to modify LLM generations in different ways. We hypothesize that this difference is due to varying writing intent mixes across \bcpwr and \wildchatwr (Figure \ref{fig-finetask-fractions}). As we see in \S\ref{sec-results}, writing intents correlate with different \proposedacronym{}s and may result in different behaviors at the dataset level when their proportions vary. Our results suggest the following implications for future work.

\textbf{\textit{Implications:}}
\begin{itemize*}
    \item Leverage implicit feedback in users' follow-ups for LLM alignment.
    \item Understand why users request revisions and leverage it for LLM alignment.
    \item Investigate users' exploration behaviors in writing sessions and examine how it can be used for better alignment.
\end{itemize*}

\begin{table}[t]
\scalebox{0.7}{
\begin{tabular}{rlc}
\toprule
Target in \bcpwr, \wildchatwr & Description & Section\\
\midrule
\proposedacronym{}2, \proposedacronym{}2 & Requesting more output. & \S\ref{sec-results-moreoutputs}\\
\proposedacronym{}3, \textsc{requests answers} & Requesting answers. & \S\ref{sec-results-followupq}\\
\proposedacronym{}4, \textsc{adds content} & Adding content. & \S\ref{sec-results-addcon}\\
\textsc{change style}, \proposedacronym{}4 & Changing style. & \S\ref{sec-results-changestyle}\\
\proposedacronym{}1, \proposedacronym{}1 & Revising requests. & \S\ref{sec-results-revising}\\
\proposedacronym{}6, \proposedacronym{}6 & Elaborating on requests. & \S\ref{sec-results-revising}\\
\bottomrule
\end{tabular}
}
\caption{Overview of \proposedacronym{}s in our regression analysis.}
\label{tab-result-themes}
\end{table}

\section{Results -- Correlating Intents and \proposedacronym{}s}
\label{sec-results}
Here we answer (RQ2): How do users' collaboration behaviors differ across writing intents? We do this by correlating writing intents and \proposedacronym{}s in regressions. When \proposedacronym{}s aren't shared across \bcpwr and \wildchatwr, we use follow-up types as targets. Table \ref{tab-result-themes} summarizes the behaviors examined in our analysis. To ensure generalization of our findings, we highlight statistically significant correlations shared across \bcpwr and \wildchatwr, and examine frequent writing intents (Figure \ref{fig-finetask-fractions}). We present the results of our mixed-methods analysis with regression coefficient plots, example conversations (truncated and rephrased) examined in qualitative analysis, and discuss the implications of our findings for future work. 

\begin{figure*}
\centering
\begin{subfigure}[c]{.45\textwidth}
   \includegraphics[trim=0.9cm 0.9cm 0cm 1.53cm, clip, width=1.05\textwidth]{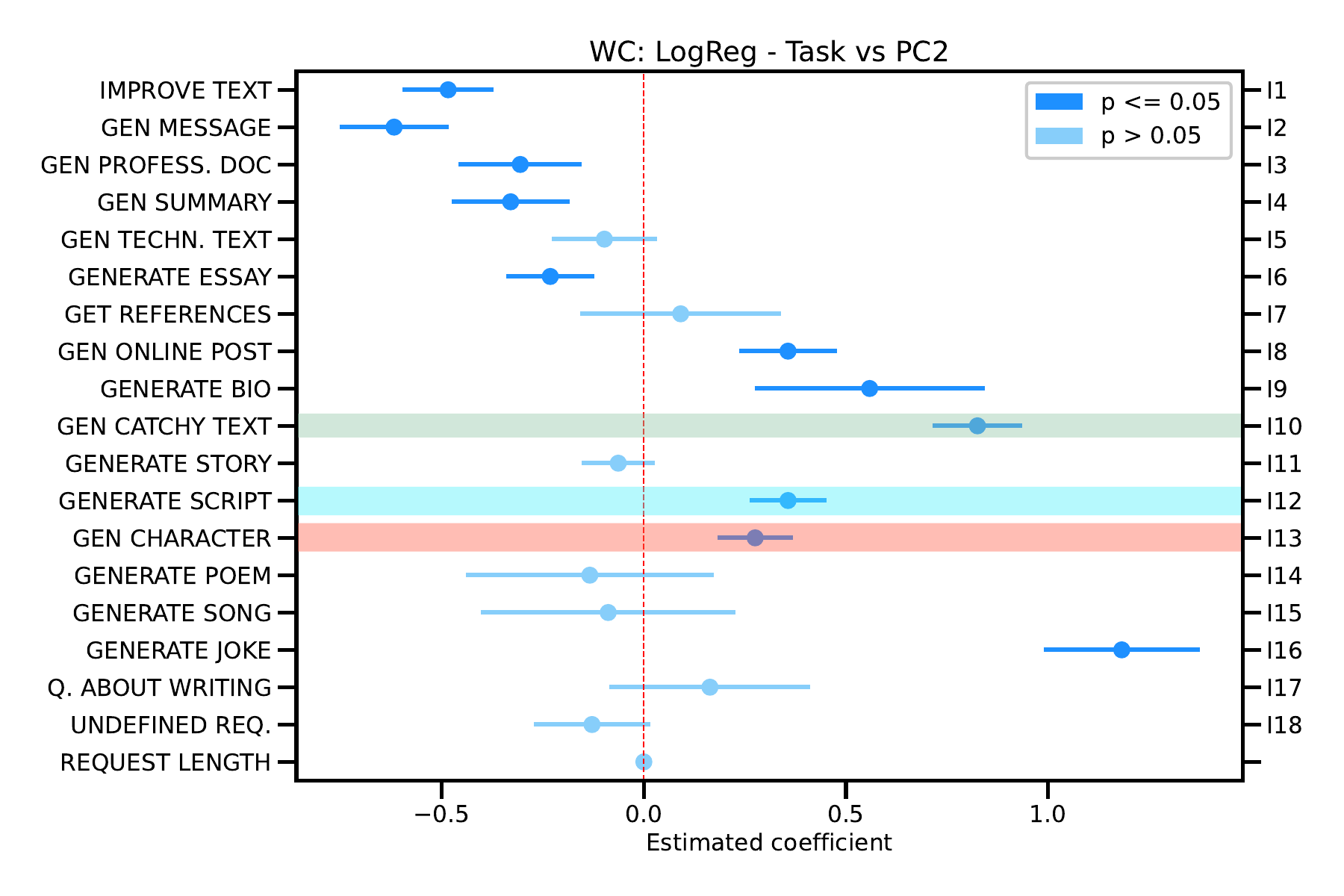}
    \caption{Logistic regression coefficients for intents vs \proposedacronym{}2 (requesting more outputs) in \wildchatwr.}
    \label{fig-coefp-wc-pc2}
\end{subfigure}~
\begin{subfigure}[c]{.5\textwidth}
\centering
    {\includegraphics[width=1.0\textwidth]{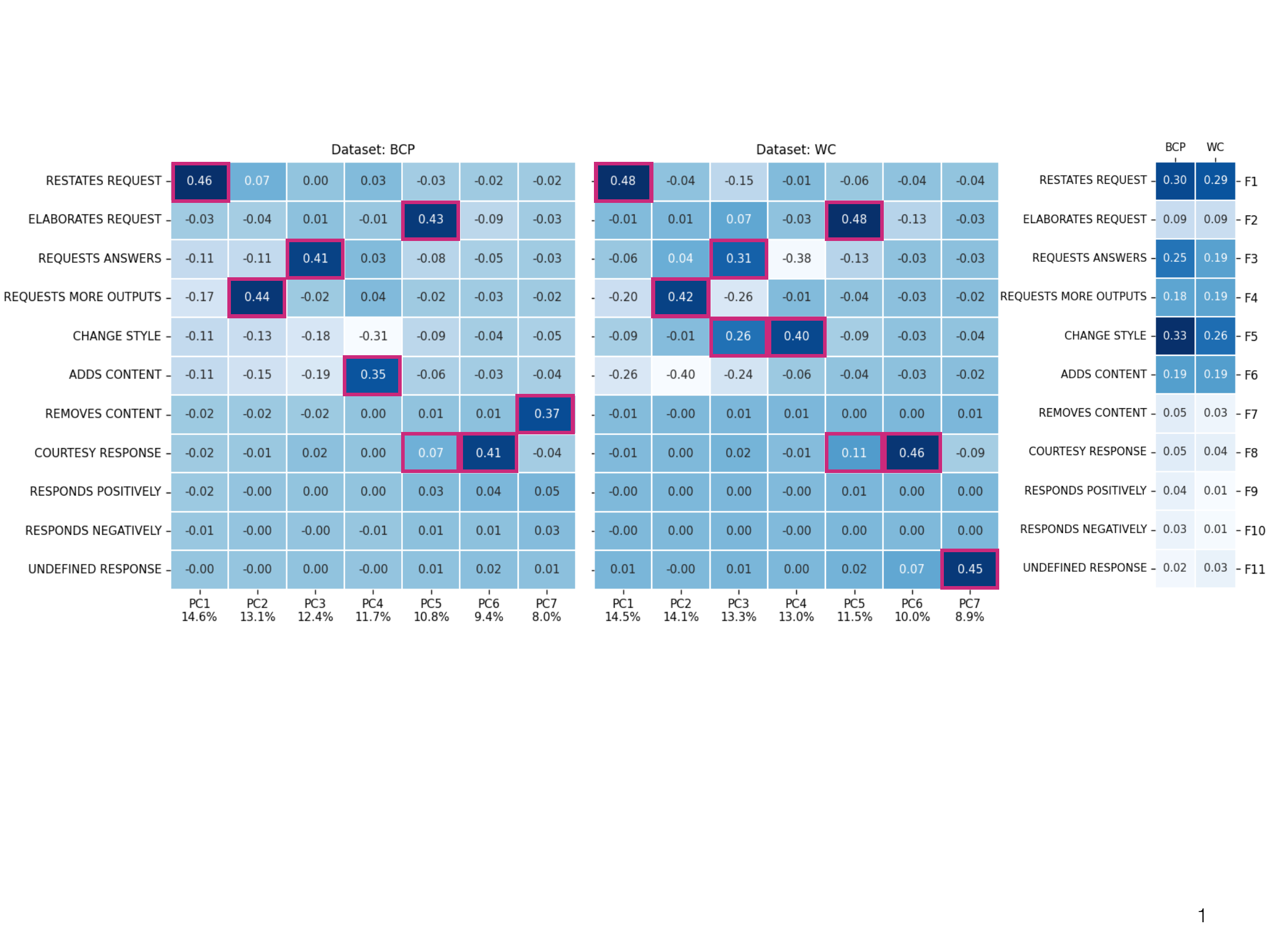}}
    \caption{\small Users engaged in intents I10, I12, and I13.}
    \label{fig-pc2-examples}
\end{subfigure}
\caption{(a) Large positive values in coefficient plots indicate strong correlations. The intents discussed in \S\ref{sec-results-moreoutputs} are highlighted in color. Coefficients for \bcpwr are plotted in Figure \ref{fig-supp-coeffplots}. (b) Example conversations from the intents highlighted in (a) -- intent and example colors are matched.}
\end{figure*}
\subsection{Requesting more outputs to brainstorm or stage long generations}
\label{sec-results-moreoutputs}
Figure \ref{fig-coefp-wc-pc2} depicts writing intents correlated with \proposedacronym{}2 where users \textsc{request more outputs}. We start by noting that \textsc{gen catchy text} (I10) shows a strong positive correlation with \proposedacronym{}2. Analysis of sessions revealed that when users aimed to generate eye-catching texts such as product names, document titles, email subjects, etc, users requested more outputs aiming to brainstorm more creative, catchy texts. Figure \ref{fig-pc2-examples} depicts examples (Ex \greenintent{1} and \greenintent{2}) from \bcpwr and \wildchatwr. This behavior finds precedent in prior work on creativity support, who note the value of diverse ideas during brainstorming \citep{frich2019cstmapping}. 
\proposedacronym{}2 may also be seen as a form of pluralistic alignment, i.e., eliciting overtone alignment \citep{sorensen2024pluralistic} -- exploring overtone alignment for brainstorming represents meaningful future work.

Next, the intents \textsc{generate script} (I12) and \textsc{gen character} (I13) show a weaker positive correlation with \proposedacronym{}2. Here, users attempted to generate media scripts  (e.g., YouTube videos) or fictional characters (Ex \redintent{4}, \neonblueintent{3}, and \neonblueintent{5} in Figure \ref{fig-pc2-examples}). Analysis of sessions revealed that while some \textsc{generate character} sessions engaged in brainstorming (Ex \redintent{4}), users primarily engaged in staged generation of long texts (Ex \neonblueintent{3} and \neonblueintent{5}). Here, users' follow-ups also varied in specificity from simply asking for more output (Ex \neonblueintent{5}) to being more specific (Ex \neonblueintent{3}). To our knowledge, this represents the first evidence of multi-turn construction of creative narratives in the wild. While prior work has explored interactive generation of long narratives through plans \citep{yao2019planwrite} interactions \citep{brahman2020cue}, and complex instructions \citep{pham2024suri}, multi-turn construction of creative narratives remains under-explored.

\textit{\textbf{Implications:}} 
\begin{itemize*}
    \item Develop and evaluate overtone-aligned LLMs for brainstorming.
    \item Develop resources and models to generate creative narratives with under-specific multi-turn interaction.
\end{itemize*}
 
\begin{figure*}
\centering
\begin{subfigure}[c]{.45\textwidth}
   \includegraphics[trim=0.9cm 0.9cm 0cm 1.53cm, clip, width=1.05\textwidth]{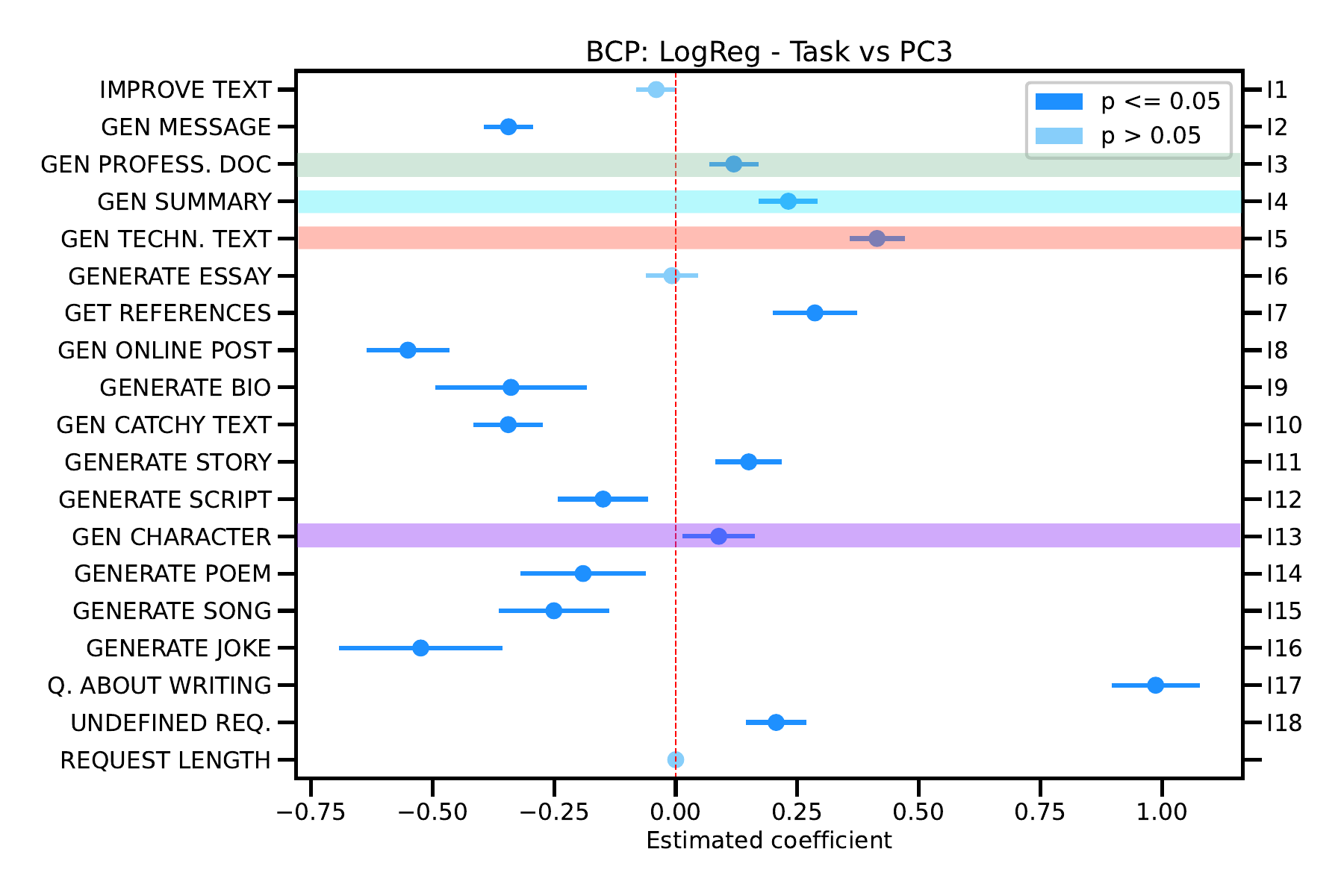}
    \caption{\small Logistic regression coefficients for intents vs \proposedacronym{}3 (requesting answers) in \bcpwr.}
    \label{fig-coefp-bcp-pc3}
\end{subfigure}~
\begin{subfigure}[c]{.5\textwidth}
\centering
    {\includegraphics[width=1.0\textwidth]{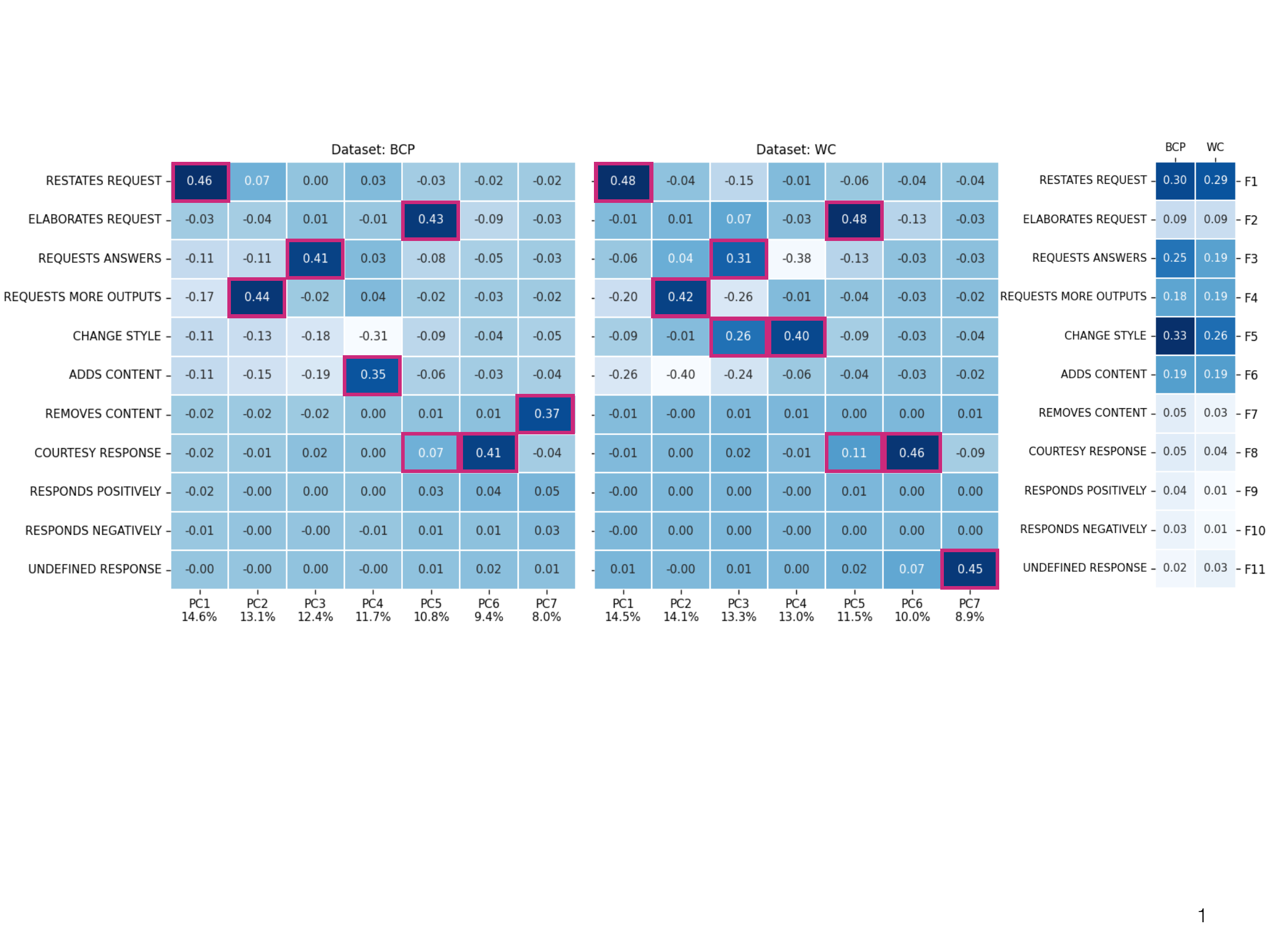}}
    \caption{\small Users engaged in intents I3, I4, I5, and I13.}
    \label{fig-pc3-examples}
\end{subfigure}
\caption{\small (a) The intents discussed in \S\ref{sec-results-followupq} are highlighted in color. Coefficients for \wildchatwr are plotted in Figure \ref{fig-supp-coeffplots}. (b) Example conversations from the intents highlighted in (a) -- intent and example colors are matched.}
\end{figure*}
\subsection{Asking follow-up questions to learn or stage long generations}
\label{sec-results-followupq}
Figure \ref{fig-coefp-bcp-pc3} shows writing intents correlated with \proposedacronym{}3, where users ask questions in response to model generations. Users tended to ask questions when they generated professional documents (I3), summaries (I4), technical texts (I5), and fictional character narratives (I13). Our analysis revealed that users' questions differed across these intents, from those asking about domain-specific knowledge and norms (Ex \greenintent{1}, \neonblueintent{2}, \greenintent{4}, \redintent{5}) to those focused on the LLMs prior generation (Ex \purpleintent{3}, \purpleintent{6}). We demonstrate this empirically in Figure \ref{fig-followup2prevgen}.

When users generated technical texts or summaries, they asked knowledge-rich questions -- needing the LLM to be grounded in domain-specific knowledge (Ex \neonblueintent{2}) or the task context (Ex \redintent{5}). %
While prior work notes that users engage in information-seeking during knowledge-rich writing \cite{shen2023beyond}  -- we provide evidence for this behavior in the wild. Information seeking during knowledge-rich writing and the use of these questions for LLM alignment remain under-studied and may be explored in future work.

We also uncover novel and emerging behaviors in users asking questions to learn about professional norms or in generating fictional narratives. When users generated professional documents such as cover letters, their questions sought to learn about professional norms and expectations (Ex \greenintent{1}, \greenintent{4}) -- needing the LLM to be grounded in these norms. While some prior work has explored writers' feedback-seeking behavior \citep{gero2023socialdyn} and LLMs' potential to provide writing feedback \citep{li2024readerprofile, weixin2024feedbackpapers}, aligning LLMs to provide feedback in writing sessions represents an emerging problem. 

Finally, when users generated characters for their stories (I13), their questions followed up on the generated stories. Analysis of sessions revealed that users sought to direct the generation of long-form fictional narratives (Ex \purpleintent{3}, \purpleintent{6}), mirroring \S\ref{sec-results-moreoutputs}. While some prior work explores question answering for fictional narratives \citep{xu2022fantastic}, a detailed understanding of users' motivations for this behavior and the use of questions to interactively build narratives remains under-explored.

\textit{\textbf{Implications:}} 
\begin{itemize*}
    \item Investigate question asking behaviors in knowledge-rich and creative writing and explore its use for session-level alignment.
    \item Evaluate and develop methods to use LLMs for providing writing feedback in high-stakes document writing.
\end{itemize*}

\subsection{Adding to generations when they lacked content known only to the user}
\label{sec-results-addcon}
\begin{figure*}
\centering
\begin{subfigure}[c]{.45\textwidth}
   \includegraphics[trim=0.9cm 0.9cm 0cm 1.53cm, clip, width=1.05\textwidth]{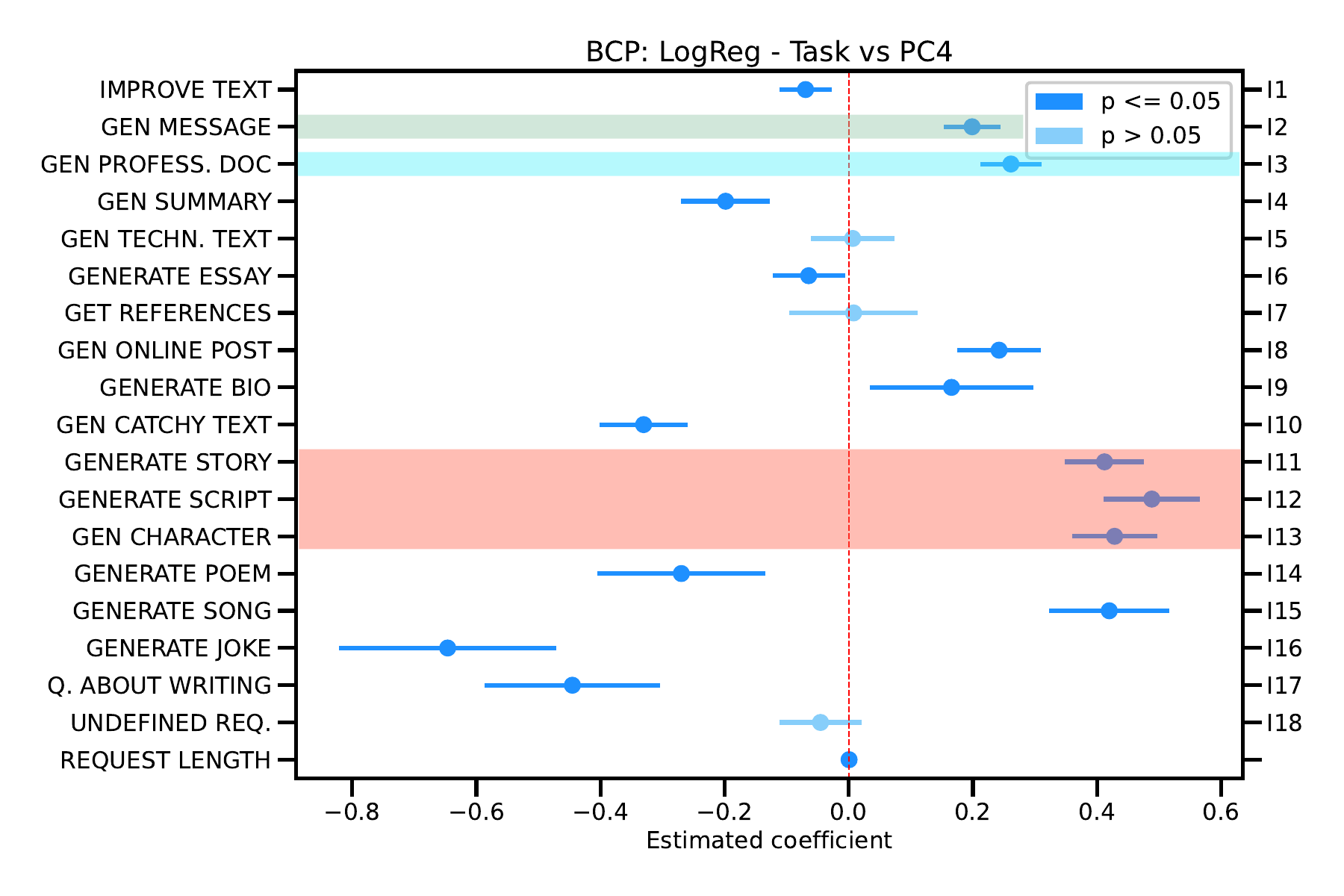}
    \caption{Logistic regression coefficients for intents vs \proposedacronym{}4 (adding content) in \bcpwr.}
    \label{fig-coefp-bcp-addcon}
\end{subfigure}~
\begin{subfigure}[c]{.5\textwidth}
\centering
    {\includegraphics[width=1.0\textwidth]{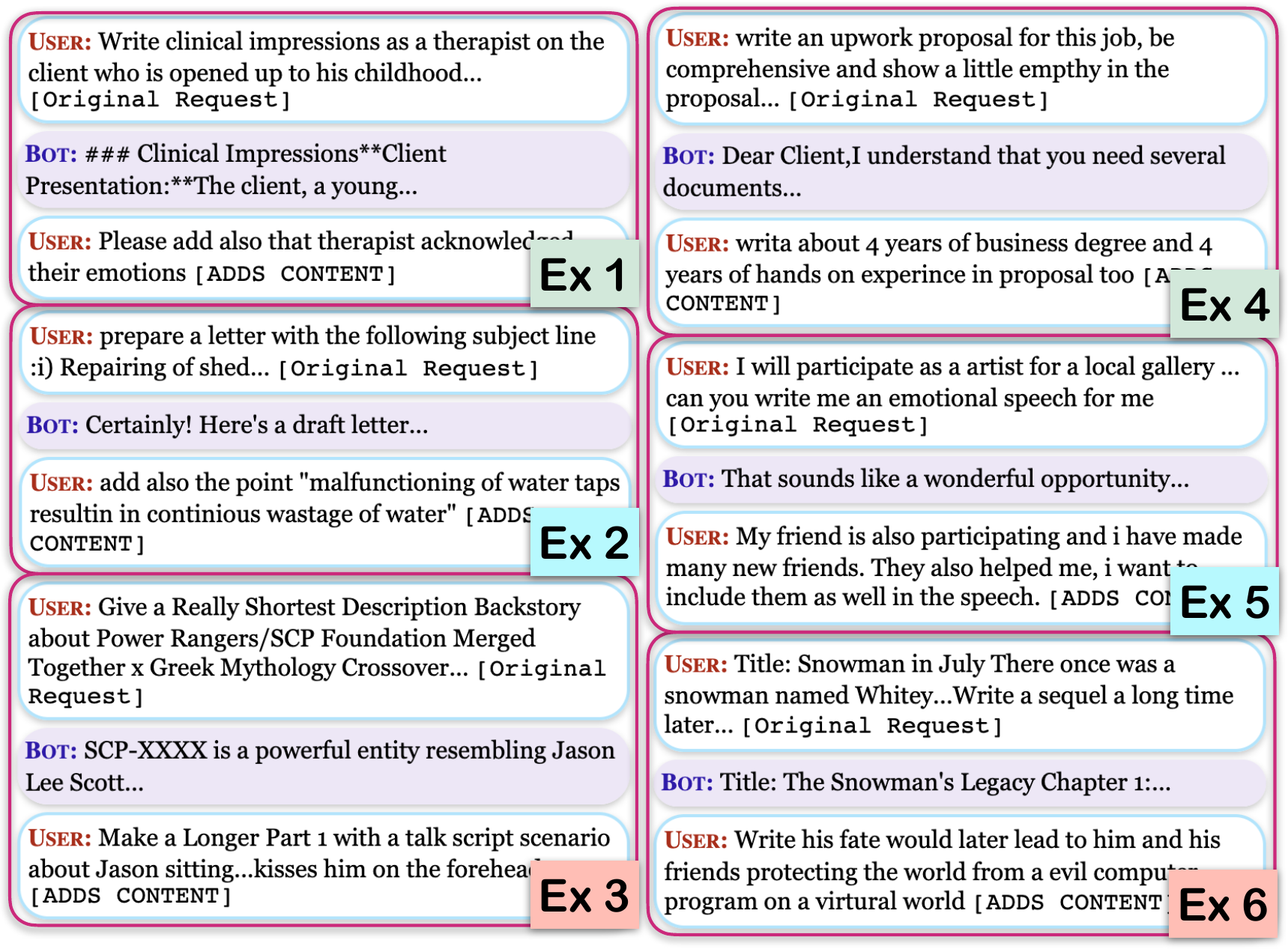}}
    \caption{Users engaged in intents I2, I3, I11-I13.}
    \label{fig-addcon-examples}
\end{subfigure}
\caption{(a) The intents discussed in \S\ref{sec-results-addcon} are highlighted in color. Coefficients for \wildchatwr are plotted in Figure \ref{fig-supp-coeffplots}. (b) Example conversations from the intents highlighted in (a) -- intent and example colors are matched.}
\end{figure*}

Figure \ref{fig-coefp-bcp-addcon} shows writing intents correlated with \proposedacronym{}4 where users add content to model generations. This behavior correlates with generating professional documents (I3), messages (I2), and creative narratives (I11-I13). 
Our analysis of sessions found that users overwhelmingly added content when model generations were missing information likely to be known only to the users. When users generated professional documents such as resumes, cover letters, clinical notes, etc.\ they added information about their personal encounters or skills (Ex \greenintent{1}, \greenintent{4}). Similarly, when they generated communicative texts such as emails, letters, or speeches, they added personal stories (Ex \neonblueintent{2}, \neonblueintent{5}). This suggests that these intents may benefit from personalization from users' historical chats or through proactive question asking to obtain missing information. While a large body of recent work has explored personalization of LLMs \citep{mysore2024pearl, magister2024way} or proactive interaction \citep{deng2025proactiveconvai}, we highlight intents where personalization or proactive interaction may be most meaningful in the wild. Finally, in generating fictional texts such as stories, scripts, or character descriptions (I11-I13), users added plots which tailored generations to their fictional visions (Ex \redintent{3}, \redintent{6}). This behavior overlaps with \proposedacronym{}s where users requested more outputs (\S\ref{sec-results-moreoutputs}), and asked questions to stage long generations (\S\ref{sec-results-followupq}), and remains understudied.

\textit{\textbf{Implications:}} 
\begin{itemize*}
    \item Develop methods to infer when long communicative texts are missing users' personal knowledge, if appropriate, obtain this information proactively, or personalize generations based on users' historical documents.
    \item Develop resources and methods to incorporate long-form feedback into creative narratives generated in multi-turn interactions. 
\end{itemize*}

In Appendix \ref{sec-results-supp} we show how users change generation style to better match readers' expectations, and find dataset-dependent trends in revising (\proposedacronym{}1) and elaborating (\proposedacronym{}5) on requests.
\section{Discussion}
\label{sec-discussion}
Taken together, our results reveal that users interact with LLMs in highly collaborative and goal-driven ways — they co-construct creative texts, probe LLMs for domain knowledge as they write, and shape text to meet long-term communicative goals. These \proposedacronym{}s highlight behaviors not captured by simpler measures of satisfaction and classification into tasks, and highlight the importance of modeling user behaviors to align LLMs with long-term user needs. Next, we synthesize the implications of our results and highlight meaningful areas of future work suggested by our results.

\textbf{Toward Session-Level Alignment from Natural Feedback.}
We showed how user follow-ups rarely included explicit positive or negative feedback in \S\ref{sec-expanalysis} and engage in open-ended collaboration behaviors in \S\ref{sec-results}. However, a majority of present work on LLM alignment remains restricted to single-turn alignment from explicit preference feedback \citep{chaudhari2024rlhf}. Our results suggest that aligning LLMs from natural feedback over multi-turn interactions presents an increasingly meaningful area of future work. Early work has begun to explore alignment with single-turn natural language feedback \citep{donyehiya2024naturalfeedback, shi2024wildfeedback} or simulated multi-turn feedback \cite{gao2024aligning, wu2025aligning}. Future work may explore building and learning from simulators that follow realistic user behaviors captured in existing log datasets or the design of reward models trained on implicit user feedback.

Further, while some \proposedacronym{}s are task-oriented and aim to directly tune LLM generations (e.g.\ changing style or adding content), others guide an exploratory session (e.g.\ requesting more outputs or answers). While task-oriented sessions have received significant attention, alignment and evaluation for exploratory sessions remain largely under-studied in the LLM-era, representing exciting future work.

\textbf{Aligning to Under-Specified Feedback and Goals.}
User follow-ups vary in their level of specificity, this is clearest in the staged generation of long texts. Here, users follow-ups increased in specificity from \textsc{requests more output} (\S\ref{sec-results-moreoutputs}), to \textsc{requests answers} (\S\ref{sec-results-followupq}), and to \textsc{adds content} (\S\ref{sec-results-addcon}). We show this empirically in Figure \ref{fig-morespecific}. This highlights a line of future work focused on LLM-alignment from under-specified natural language feedback.

Most current work assumes user interactions (ORs and \proposedacronym{}s) to be fully specified. While a sizable body of work has focused on uncertainty estimation for LLM outputs \citep{geng2024survey}, limited work has explored uncertainty in user interactions. This early work has sought to resolve under-specificity through task context \cite{malaviya2024contextualized, sarkar2025promptrevision}, proactive question asking \cite{kobalczyk2025active, li2025aligning}, and coverage-based methods \cite{hou2024decomposing}. Exploring these and other strategies is a valuable direction for future work.

\textbf{LLMs as Learning Partners: Aligning for User Growth.}
Across a writing intent and \proposedacronym{}s, we find users to leverage LLMs not merely as task assistants, but as partners in their learning and development. Users brainstormed with models to generate more compelling or creative text (\S\ref{sec-results-moreoutputs}), asked questions to understand professional norms while drafting documents (\S\ref{sec-results-followupq}), requested feedback to improve fictional narratives (\S\ref{sec-results-followupq}), and iterated on text to better suit their audiences (\S\ref{sec-results-changestyle}). These behaviors reflect a desire not only to produce better outputs, but to use LLMs for longer-term learning.

Our findings suggest that aligning LLMs to support users' growth and learning represents a compelling and underexplored design goal. Recent work has echoed this vision \citep{collins2024building}, and prototype systems are beginning to emerge \citep{li2024readerprofile, chamoun2024automated}. However, realizing this potential will require addressing critical challenges: mitigating the risk of cultural bias in ``coaching,'' preventing homogenization of users' voice and ideas \citep{agarwal2025homogenize, drosos2025makes}, and designing interactions that preserve the ``productive struggle'' known to facilitate deeper learning \citep{rus2025llmslearning}.

Ultimately, building LLM systems for learning will demand rich interdisciplinary work, drawing on learning sciences, HCI, and NLP to understand how LLM systems can guide and challenge users, rather than merely serve their short-term needs.
\section{Limitations}
\label{sec-limitations}
In this paper\footnote{LLM assistants were used to improve the phrasing of some sentences in this paper and fix grammatical errors.}, we analyzed a large sample of Bing Copilot and WildChat user-LLM interaction logs spanning seven and thirteen months of global usage. We focused on user-LLM collaboration behaviors in English writing sessions. Our analysis took a mixed-methods approach, using PCA to identify prototypical human-AI collaboration behaviors (\proposedacronym{}s) and regression analysis to identify statistically significant correlations between writing intents and \proposedacronym{}s. We paired this with lightweight qualitative analysis to gain further insight into automatically highlighted behaviors. We ensure the reliability of our analysis through manual validation of all automated components. We contribute the first understanding of in-the-wild user-LLM collaboration behaviors and discuss their implications for LLM alignment. Now, we highlight some drawbacks of our analysis and its broader impact.

\textbf{Data and Analysis.} Our analysis was limited to English sessions. Therefore, it is likely that some collaboration behaviors specific to non-English users were missed in our analysis. We made this choice to ensure the reliability of LLM classifiers and to ensure that log text was intelligible to all authors during annotation and analysis. Extending our analysis to non-English logs represents a meaningful area of future work. Next, our analysis (\S\ref{sec-method}) does not model the order of follow-ups in a session and uses linear PCA to identify \proposedacronym{}s. We rely on simpler modeling to ensure the interpretability of results in a first analysis, future work may explore richer modeling of follow-ups to discover more complex \proposedacronym{}s. Finally, we rely on a light-weight qualitative analysis to understand users' goals in correlated intent-\proposedacronym{} pairs. Future work may explore these behaviors in controlled human-centered studies with direct access to the actual users.

\textbf{Broader Impacts.} We highlight LLM alignment from user-LLM interactions to be a meaningful area of future work in \S\ref{sec-results}. However, we note that learning from implicit interactions may pose safety risks if user interactions are adversarial (e.g. \citet{zhao2024challenges}) or pose privacy risks (e.g.\ \citet{xin2024a}) when implicit interactions encode private user information. This may result in user interactions being leaked from aligned LLMs. Therefore, ensuring safety and privacy must be an important aspect of learning from implicit interactions. 

\bibliography{ipt-bib}

\appendix
\clearpage

\section{Extended Related Work}
\label{sec-relatedwork-supp}
Having discussed prior work on user-LLM log analysis in \S\ref{sec-relatedwork} we discuss additional related work here.

\textbf{Measuring user satisfaction.} Our analysis of users' follow-up behaviors also ties it to prior work on predicting binary signals of dis/satisfaction from conversations \citep{lin2024interpretablesat, biyani2024rubicon, mahato2024exploring}. In our work, we go beyond binary notions of dis/satisfaction and provide a richer characterization of users' collaboration behaviors. Further, we find explicit satisfaction signals to be rare and highlight a need to understand and use implicit behaviors for LLM alignment (see \S\ref{sec-expanalysis}). Concurrent analysis on Bing Copilot and WildChat-1M logs is also relevant to our work. Here, \citet{palta2025expertisesat} correlates the difference between expertise in human and LLM utterances with binary metrics of user satisfaction. And \citet{brigham2024developingstory} conduct a small-scale analysis of journalistic writing sessions in WildChat-1M and discuss its implications for responsible AI use in journalism. While these works differ in focus, together with our work, they paint a more complete picture of human-LLM interactions in the wild.


\textbf{Human-centered Studies on Writing Assistance} A sizable body of work in HCI has focused on writing assistance. Recent surveys review this body of work more completely \citep{lee2024desigspace, zhao2025making}. Prior work examining interactions with LLM writing assistants is most relevant to our work. In contrast to user-LLM log analysis, this research has had more direct interaction with users -- enabling studies to probe user perceptions, behaviors, and impacts more deeply while forgoing an understanding of in-the-wild user behaviors outside study settings. Broadly, this work has focused on the impacts of LLM-assisted writing \citep{hancock2020aimcagenda}, studied user-LLM co-creation behaviors, and designed novel interaction paradigms \citep{buschek2024collage}. Work on impacts has examined aspects such as writers perceived ownership over texts co-created with personalized LLMs \citep{draxler2024aighostwriter, hwang2024it80me20}, convergence toward normative stances, stereotypes, and styles in LLM assisted writing \citep{ceci2024autocompletebias, huang2024gptgender, agarwal2025homogenize}, and the perception of fit between LLM outputs and tasks contexts \citep{fu2024texttoself, bhattacharjee2024chatpreference}, among others. Studies focusing on the dynamics of human-AI co-creation examine the differences in how writers seek feedback from humans vs LLMs \citep{gero2023socialdyn}, or how they seek assistance at different stages of writing \citep{wan2024secondmind, chakrabarty2024creativitysupportagelarge}. Still others have examined the impact of specific interaction behaviors on productivity and decision making \citep{bhat2023nextphrase, dhillon2024scaffolding}. 
Our analysis complements this line of work by supporting laboratory findings through large-scale analysis where applicable, and uncovers novel user behaviors likely to be missed because of smaller samples or constrained study designs.

\section{Analysis Setup - Details}
\label{sec-analysis-setup-supp}
In \S\ref{sec-analysis-setup} we presented an overview of our writing session datasets \bcpwr and \wildchatwr filtered from Bing Copilot and WildChat-1M logs. Both datasets are obtained after two broad stages of filtering, we describe this filtering here:
\begin{itemize}[nolistsep, noitemsep]
    \item Appendix \ref{sec-filteringwriting-supp}: Preliminary filtering, i.e. Bing Copilot $\rightarrow$ \bcpall and WildChat-1M $\rightarrow$ \wildchatall.
    \item Appendix \ref{sec-taskclf-supp}: Writing session filtering i.e. \bcpall $\rightarrow$ \bcpwr and \wildchatall $\rightarrow$ \wildchatwr.
    \item Appendix \ref{sec-taskclf-validation-supp}: Validating Task Classifiers used to identify writing sessions.
\end{itemize}

\subsection{Dataset Description and Preliminary Filtering}
\label{sec-filteringwriting-supp}
\textbf{Bing Copilot $\rightarrow$ \bcpall}. We start with a daily random sample of Bing Copilot logs from a seven-month period spanning 1 April to 31 October 2024, amounting to 20.5M conversational sessions. GPT-4 powered Bing Copilot during this period. We perform an initial filtering based on logged metadata. Since we aim to study users' follow-up behaviors, we retain sessions with 2 or more user utterances, sessions that are likely to be in English (based on the users system language) to ensure familiarity of data to all authors, and sessions conducted on a personal computer -- this results in 3.6M sessions. We choose to retain sessions on personal computers to limit the impact of behavioral differences across devices, e.g., prior work notes limited diversity of queries in search on mobile phones compared to personal computers \citep{kamvar2006googlemobile}. 

Next, we perform a second stage of language filtering based on the first 1500 characters (about 300 words) of a session text using a Language ID model. \footnote{FastText Language ID model: \href{https://fasttext.cc/docs/en/language-identification.html}{\texttt{lid.176.bin}}} Metadata-based filtering alone proved to be insufficient to filter non-English sessions. We retain 2.8M English sessions, and we refer to this as \bcpall. This dataset was labeled with a coarse Task Classifier (\coarsetaskclf) to identify writing sessions.
Note that all our Bing Copilot data was scrubbed for personally identifying information, held on secure servers, labeled with privacy-compliant LLM deployments, and our analysis was conducted in accordance with organizational privacy policies and data handling practices.
\begin{figure}
  \includegraphics[width=\textwidth]{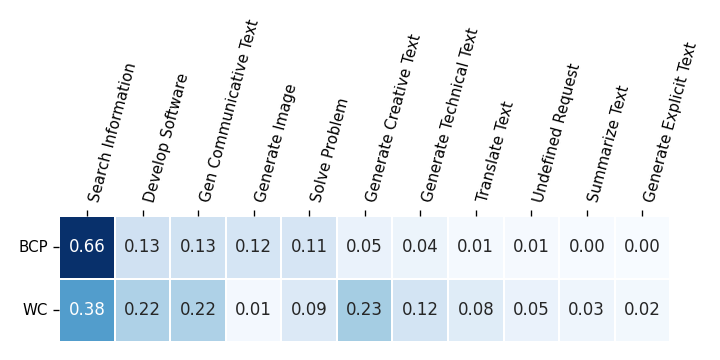}
  \caption{The proportion of coarse tasks in \bcpall (2.8M sessions) and \wildchatall (160k sessions). The sessions are in English and likely to have at least 1 user follow-up. A session can have multiple coarse tasks. On average, \bcpall and \wildchatall had 1.25 and 1.43 tasks per session.}
  \label{fig-coarsetask-fractions}
\end{figure}

\textbf{WildChat-1M $\rightarrow$ \wildchatall}. We start with the public\footnote{\href{https://huggingface.co/datasets/allenai/WildChat-1M}{allenai/WildChat-1M}. Accessed 12 December 2024.} WildChat-1M dataset available under an AI2 ImpACT license. \citet{zhao2024wildchat} gathered this data over thirteen months from 9 April 2023 to 1 May 2024 in a HuggingFace Hub space. The sessions contain interactions with GPT-4 and GPT-3.5-Turbo. Of the 840k publicly available sessions, we retain sessions labeled as English in the metadata and with 2 or more user utterances -- this results in 160k sessions, we refer to this as \wildchatall. Then \wildchatall sessions are labeled with \coarsetaskclf to identify writing sessions. 

\textbf{Dataset Characteristics} In Table \ref{tab-exploratory-countries} we show the number of users and countries of origin for \bcpall and \wildchatall. In Figure \ref{fig-coarsetask-fractions} we show the proportion of tasks identified by \coarsetaskclf before filtering them down to writing sessions alone. We see that Bing Copilot sessions are focused on search (``Search Information''), whereas WildChat has a more even proportion of tasks. A likely explanation is that Bing Copilot was available alongside the Bing search engine, whereas WildChat was framed as a general-purpose conversational assistant in a research study. Further, \coarsetaskclf, identified 1.25 and 1.43 tasks per session on average in \bcpall and \wildchatall, respectively. The median was 1 task -- indicating that most sessions focus on a single coarse task.
\begin{table}[t]
    \scalebox{0.83}{
        \begin{tabular}{rccc}
        \toprule
        & Utterances & Original & Follow\\
        & & requests & ups\\
        \midrule
        \bcpwr &   $7.5$~/~$6$  &  $1.6$~/~$1$  &  $2.1$~/~$2$   \\
        \wildchatwr  &  $7.1$~/~$6$ & $1.6$~/~$1$ & $1.9$~/~$1$ \\
        \bottomrule
        \end{tabular}
    }
    \caption{Session level characteristics for our datasets (mean~/~median). We classify session utterances from the user into Original Requests and Follow ups, where ORs may have 1 or more fine-grained intents.}
    \label{tab-exploratory-session-stats}
\end{table}


\subsection{Identifying Writing Sessions}
\label{sec-taskclf-supp}
\textbf{Task Classifier.} We follow prior work and conceptualize writing sessions to be the ones where users translate their thoughts into writing \cite{lee2024desigspace}. We interpret this broadly and treat sessions where users write or re-write, whole or parts of texts to be a writing session. We implement this in GPT-4o based \coarsetaskclf and retain sessions labeled as generating communicative, creative, technical, and summarizing text as writing tasks -- Figure \ref{fig-coarsetask-fractions} depicts their proportions.

Prompt \ref{fig-coarsetaskclf-prompt} was used for \coarsetaskclf. We generate our task labels and label definitions by collating tasks identified in \citet{suri2024binganalysis}, and through additional manual analysis of 200 \bcpall conversations. We ensure the label coverage by manually applying them to 200 sessions in \wildchatall. In both data sets, \coarsetaskclf inputs the first 10 utterances to predict a session's coarse task. Given the larger size of \bcpall and the expense (time and dollar amounts) of GPT-4o calls, we trained an embedding-based scalable classifier on labels obtained from GPT-4o for 100k \bcpall sessions. Our scalable classifier used a frozen \texttt{
E5-large-v2} \cite{wang2024e5embeddings} model for embedding the conversation text and fine-tuned logistic regression classifiers in a one-vs-all setup to obtain a \coarsetaskclf for \bcpall. Since \wildchatall is a smaller dataset, we used GPT-4o directly for it.

\textbf{\bcpall $\rightarrow$ \bcpwr} To obtain writing sessions, we label \bcpall sessions with \coarsetaskclf and retain sessions labeled with the labels Gen Communicative Text, Gen Creative Text, Gen Technical Text, and Summarize Text as writing sessions. Out of 2.8M sessions, we identify 500k as writing sessions. These made up 17-19\% sessions in each month. Finally, we randomly subsample 250k of 500k, resulting in \bcpwr. Our sub-sampling was done to lower the costs of subsequent analysis, which relies on LLM classifiers while retaining a large enough sample for analysis.

\textbf{\wildchatall $\rightarrow$ \wildchatwr} Out of 160k sessions, 68k (42\%) were identified as writing sessions by \coarsetaskclf to form \wildchatwr. We follow the same filtering process as for \bcpall.

Session length characteristics of \bcpwr and \wildchatwr are shown in Table \ref{tab-exploratory-session-stats}. We see that sessions had 2 follow-ups on average, enabling our analysis to focus on collaboration behaviors.

\begin{table}[t]
\scalebox{0.82}{
\begin{tabular}{lcc}
\toprule
    & \multicolumn{2}{c}{Coarse Tasks} \\
    \cmidrule(lr){2-3}
    & Acc. ($\%$)      & Agr. ($\kappa$)\\
\midrule
\bcpall & 89.09 & 1.00\\
\wildchatall  & 80.00 & 0.87\\
\midrule
    & \multicolumn{2}{c}{Writing Tasks} \\
\midrule
\bcpall & 85.71 & 1.00\\
\wildchatall  & 91.07 & 0.87\\
\bottomrule
\end{tabular}
}
\caption{Accuracy (Acc.) of GPT-4o based Task Classifier (\coarsetaskclf) in \bcpall and \wildchatall. Prompt \ref{fig-coarsetaskclf-prompt} contains label definitions for the eleven coarse task classes. Four of the eleven classes are treated as writing sessions -- accuracy and agreement for these classes alone are also included above. Agreements (Agr.) are computed from judgments made by two independent annotators.}
\label{tab-coarsetaskclf-accuracy}
\end{table}

\subsection{Validating Task Classifiers}
\label{sec-taskclf-validation-supp}
\textbf{Setup.} Table \ref{tab-coarsetaskclf-accuracy} presents the results of manually validating \coarsetaskclf. Our validation aims to establish the accuracy of identifying coarse tasks so that subsequent analysis is not affected by noisy predictions. Two annotators judged the correctness of \coarsetaskclf predictions for 55 sessions from \bcpall and \wildchatall -- 110 sessions in all. We sample an equal number of sessions for each label. The two annotators' judgments were used to compute agreements with Cohen's Kappa, $\kappa$. In Table \ref{tab-coarsetaskclf-accuracy} we report metrics across all labels and for the labels considered as writing (i.e.\ the labels Gen Communicative Text, Gen Creative Text, Gen Technical Text, and Summarize Text). 

\textbf{Results.} From Table \ref{tab-coarsetaskclf-accuracy} we see that \coarsetaskclf is able to identify writing sessions in \bcpall and \wildchatall with a sufficiently high accuracy and substantial agreement. We also note that \bcpall has a slightly lower accuracy than \wildchatall for identifying writing tasks. We attribute this to the difference in task frequencies in the two datasets (see Figure \ref{fig-coarsetask-fractions}). Bing Copilot was used most frequently for search, and most errors occurred because of mislabeling search sessions as a different task. On the other hand, WildChat was used in nearly equal proportion for various tasks, with errors resulting from different types of writing tasks confused for each other (e.g., communicative vs creative). Since different writing tasks are collapsed into one writing class, the accuracy of identifying writing sessions remains high.

\section{Log Analysis with \proposedacronym{}s - Details}
\label{sec-method-supp}

In \S\ref{sec-method} we describe our analysis method focused on identifying \proposedacronym{}s and on analyzing how \proposedacronym{}s vary across writing intents (Figure \ref{fig-analysis-overview}). We provide the details of the methods here:
\begin{itemize}[nolistsep, noitemsep]
    \item Appendix \ref{sec-method-identifypref-supp}: Details for the Follow-up Classifier and Identifying \proposedacronym{}s with PCA.
    \item Appendix \ref{sec-method-correlatepref-supp}: Details for with Intent Classifier, Regression Analysis, and Qualitative Analysis.
    \item Appendix \ref{sec-method-validation-supp}: Details about manually validating the Follow-up and Intent Classifier.
\end{itemize}

\subsection{Identifying \proposedacronym{}s}
\label{sec-method-identifypref-supp}
We identify \proposedacronym{}s by first classifying users' follow-up utterances into high-level follow-up types and clustering these into \proposedacronym{}s using PCA. 

\textbf{Follow-up Classifier.} We use a GPT-4o based follow-up classifier (\followupclf) to classify user utterances in \bcpwr and \wildchatwr into original requests or one of eleven follow-up types $\mathcal{F}$ (Table \ref{tab-follow-up-descr}). Table \ref{tab-follow-up-examples} shows example follow-up utterances for each label. 

Prompt \ref{fig-fgfollowup-prompt} was used for \followupclf and contains labels and their definitions. Our follow-up type labels were derived by clustering user utterances embedded with a pretrained language model and manually generating labels for each cluster. We iteratively repeat this process until no new labels are generated. Then the labels were merged to remove redundancies. We confirm their coverage by manually applying the labels to 100 randomly sampled \bcpwr sessions. Our labels (Table \ref{tab-follow-up-descr}) closely mirror labels generated in prior conversational interaction datasets \citep{qu2018msdialog} -- indicating their generality. We validate the accuracy of \followupclf through manual annotation in Appendix \ref{sec-method-validation-supp}.

\textbf{\proposedacronym{}s with PCA.} After labeling user utterances with \followupclf we identify \proposedacronym{}s. Given a dataset of sessions $\mathcal{S}$, we represent each session $S\in\mathcal{S}$ using a ``\texttt{tf-idf}'' representation of follow-up types, $\mathbf{F} \in \mathbb{R}^{|\mathcal{S}|\times|\mathcal{F}|}$. Each feature vector per session in $\mathbf{F}$ contains the follow-up type counts normalized by the total number of user utterances in the session (``\texttt{tf}''). These frequencies are then multiplied by the log inverse of the follow-up frequency in the entire dataset (``\texttt{idf}''). This ensures that more frequent follow-ups don't dominate the variance of our dataset. 

Next, we run PCA on $\mathbf{F}$ and treat each identified principal component as a \proposedacronym{}. PCA is a linear dimensionality reduction technique which maps $\mathbf{F}$ to a new space of variables as $\mathbf{P} = \mathbf{F}\mathbf{W}$. The dimensions of $\mathbf{P}$ are uncorrelated with each other, and $\mathbf{W}$ ensures that the variance of $\mathbf{F}$ is retained in $\mathbf{P}$ \citep{bengio2013replearn}. 
Further, the dimensions of $\mathbf{P}$ capture decreasing amounts of the variance in $\mathbf{F}$. In our work, we retain the first $l$ ($< |\mathcal{F}|$) dimensions of $\mathbf{P}$ that explains 80-85\% variance in $\mathbf{F}$ while rejecting the rest as noise, therefore, $\mathbf{P}_l \in \mathbb{R}^{|\mathcal{S}|\times l}$. Because each dimension of $\mathbf{P}_l$ represents a mutually co-occurring set of follow-ups, they segment $\mathcal{S}$ into sessions with a consistent \proposedacronym{}. We visualize $\mathbf{W}_l$, referred to as the ``loading matrix'', in \S\ref{sec-expanalysis} to show the correlations between follow-up types and \proposedacronym{}s.

\subsection{Correlating intents and \proposedacronym{}s}
\label{sec-method-correlatepref-supp}

\textbf{Intent Classifier.} To correlate writing intents with \proposedacronym{}s, we classify ‘Original Requests’ identified by \followupclf into fine-grained writing intents (Table \ref{tab-fgtype-descr}) using a GPT-4o based classifier (\fineintclf). We set up \fineintclf as a multi-label classifier since requests can express more than one intent. 

Prompt \ref{fig-fgintents-prompt} was used for \fineintclf and contains labels and their definitions. Mirroring label generation for follow-ups, we derive writing intents iteratively. We cluster user utterances in \bcpwr and manually generate intent labels per cluster while merging redundant intents. We repeat this procedure until new labels aren't generated. We manually apply the labels to 200 randomly sampled \bcpwr sessions to ensure their coverage. Table \ref{tab-fgtype-examples} shows example original requests for each writing intent. We validate the accuracy of \fineintclf through manual annotation in Appendix \ref{sec-method-validation-supp}. 

\textbf{Regression Analysis.} We run logistic regression analysis\footnote{\href{https://www.statsmodels.org/stable/generated/statsmodels.discrete.discrete_model.Logit.html}{\texttt{statsmodels.Logit}}} by treating fine-grained intents of a session as predictors and the sessions \proposedacronym{} as a target variable. We rely on logistic regressions due to their ease of interpretation and follow a large body of prior work  \citep{gujarati2021essentials}. 
In our regression, we represent the fine-grained intents identified by \fineintclf as one-hot features, $\mathbf{I}$, of the sessions -- these serve as predictors. Next, we obtain the sessions membership in $l$ different \proposedacronym{}s, $\mathbf{M}_l$, by thresholding the columns of $\mathbf{P}_l$. Then, we correlate the intents $\mathbf{I}$ with \proposedacronym{}s of $\mathbf{M}_l$. This results in $l$ independent logistic regressions: $\texttt{sigmoid}(b + \mathbf{I}\mathbf{c})$. In thresholding $\mathbf{P}_l$, we select a score to retain 15-20\% of the sessions with the highest scores per dimension in $\mathbf{P}_l$. Our thresholds were selected to ensure the accuracy of our logistic regressions. In our results, we examine the coefficients $\mathbf{c}$ -- correlating the intents with each one of the $l$ \proposedacronym{}s. Finally, note that the presence of a follow-up type in a session can be used directly as a binary target variable instead of PCA-based \proposedacronym{}s -- we pursue both paths in our analysis.
\begin{figure}
  \includegraphics[width=\textwidth]{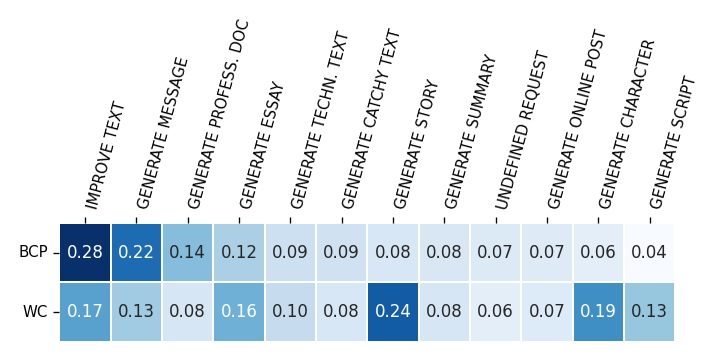}
  \caption{Frequencies of writing intents in \bcpwr and \wildchatwr identified by multi-label \fineintclf. From the top-3 intents, we see that Bing Copilot was used for creating communication or professional texts, while WildChat was used for creative texts such as stories, scripts, or characters. Intents occurring in fewer than 3\% sessions are excluded for space.}
  \label{fig-finetask-fractions}
\end{figure}

Note that we also use the length of original requests as predictors and find  that they are uncorrelated with \proposedacronym{}s (\S\ref{sec-results}). Similarly, early analysis also found the deployment LLM be uncorrelated with \proposedacronym{}s -- likely due to \proposedacronym{}s capturing high-level behaviors. Therefore they were dropped as features in subsequent analysis. To indicate the goodness of classifier fit, Table \ref{tab-result-logreg-acc} presents the accuracy of classifiers used to identify correlations between writing intents and \proposedacronym{}s/follow-up types and shows significantly better accuracy than a random classifier (50\% accuracy).

\textbf{Qualitative Analysis.} Our automatic approach highlights correlated writing intents and \proposedacronym{}s. To gain a deeper understanding of user behaviors in a correlated intent-\proposedacronym{} pair, two authors conducted a qualitative analysis of the pairs, which showed statistically significant correlations and were repeated in \bcpwr and \wildchatwr. For each pair, both authors independently examined the top 300 conversations with the highest principal component scores, aiming to understand the user goals for engaging in a \proposedacronym{}. Then both authors met and resolved the differences in their understanding. In our analysis, the two authors differed in the interpretation of one of the six intent-\proposedacronym{}s presented in our results. Our analysis was conducted to overcome the lack of access to Bing Copilot or WildChat users who could be probed about their intents and behaviors -- a fundamental challenge in all log-based studies \citep{dumais2014loganalysis}.
\begin{table}[t]
\scalebox{0.7}{
\begin{tabular}{rlc}
\toprule
Target in \bcpwr, \wildchatwr & Description & Accuracy (\%)\\
& & \bcpwr, \wildchatwr\\
\midrule
\proposedacronym{}2, \proposedacronym{}2 & Requesting more output. & 79, 79\\
\proposedacronym{}3, \textsc{req. ans.} & Requesting answers. & 77, 74 \\
\proposedacronym{}4, \textsc{adds con.} & Adding content. & 76, 74\\
\textsc{ch. sty.}, \proposedacronym{}4 & Changing style. & 61, 78 \\
\proposedacronym{}1, \proposedacronym{}1 & Revising requests. & 73, 74\\
\proposedacronym{}6, \proposedacronym{}6 & Elaborating on requests. & 84, 84\\
\bottomrule
\end{tabular}
}
\caption{The accuracies for logistic regressions used to find correlations between writing intents and \proposedacronym{}s or follow-up types in our regression analysis.}
\label{tab-result-logreg-acc}
\end{table}

\subsection{Validating Follow-up and Intent Classifiers} 
\label{sec-method-validation-supp}

\textbf{Setup.} Table \ref{tab-llmclf-accuracy} presents the results of manually validating \followupclf and \fineintclf. Similar to \coarsetaskclf, our annotation validates the accuracy of \followupclf and \fineintclf to establish that the subsequent analysis is error-free. Validation was conducted by two co-authors serving as annotators, they manually judged the correctness of \followupclf predictions for follow-ups and \fineintclf predictions for original requests. Both annotators judged the correctness of the labels independently and on the basis of a shared annotation guideline. Predictions were considered incorrect if a different label was applicable. The two annotators' judgments were used to compute agreements with Cohen's Kappa, $\kappa$. We constructed our validation data by selecting follow-ups from 110 sessions and original requests from a different set of 110 sessions -- 220 sessions in all. We ensured that every follow-up and writing intent label was uniformly present in the validation data. \bcpwr and \wildchatwr had 110 sessions each in the validation data. For the multi-label \fineintclf annotators independently judged the correctness of all predicted labels. 

\textbf{Results.} From Table \ref{tab-llmclf-accuracy} we note that the follow-up and writing intent classifiers have a reasonably high accuracy and substantial agreement between the annotators. Our annotators noted that most writing intent errors were the result of hard-to-distinguish boundaries between creative writing tasks (i.e.\ \textsc{generate story}, \textsc{generate character}, \textsc{generate script}). Similarly, follow-up type errors were the result of hard-to-distinguish boundaries between different ways to update a generation (i.e.\ \textsc{change style}, \textsc{add content}, \textsc{remove content}). 

\begin{table}[t]
\scalebox{0.82}{
\begin{tabular}{lcccc}
\toprule
    & \multicolumn{2}{c}{Follow-ups} & \multicolumn{2}{c}{Intents} \\
    \cmidrule(lr){2-3}\cmidrule(lr){4-5}
    & Acc. ($\%$)      & Agr. ($\kappa$)     & Acc. ($\%$)      & Agr. ($\kappa$)\\
\midrule
\bcpwr & 79.09 & 0.78 & 81.58 & 0.79\\
\wildchatwr  & 84.55 & 0.74 & 78.70 & 0.82\\
\bottomrule
\end{tabular}
}
\caption{Accuracy (Acc.) of GPT-4o classifiers for labeling follow-up types (\followupclf) and writing intents (\fineintclf) in \bcpwr and \wildchatwr. Agreements (Agr.) are computed from judgments made by two independent annotators.}
\label{tab-llmclf-accuracy}
\end{table}
\section{Extended Results}
\label{sec-results-supp}
\begin{figure*}
\centering
   \includegraphics[width=0.7\textwidth]{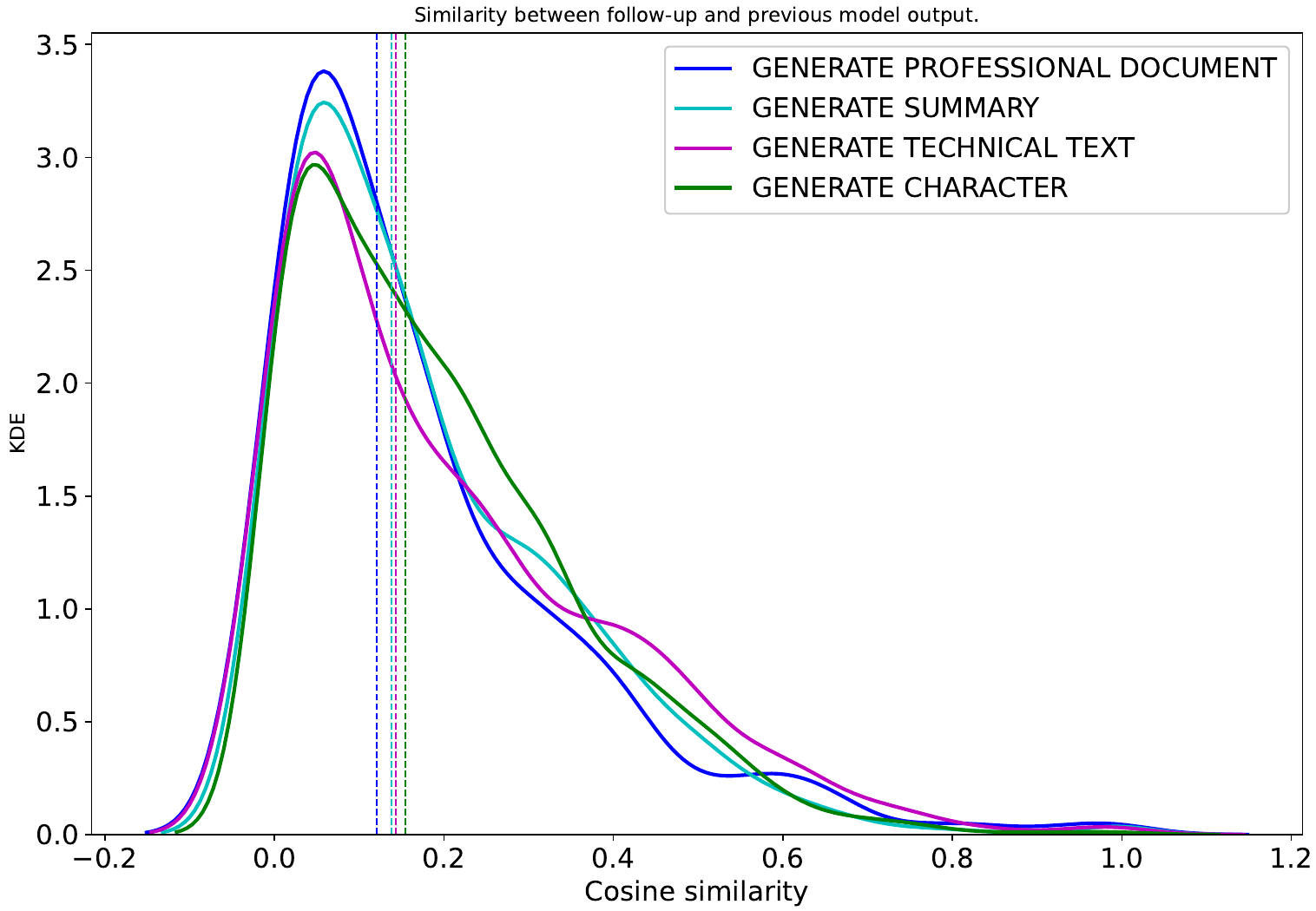}
    \caption{Increasing similarity between the follow-up utterance and the previous model generation in \wildchatwr for writing intents correlated with \proposedacronym{}3 (asking follow-up questions). We discuss this in \S\ref{sec-results-followupq}. Similarities are increase from users co-creating professional documents, to summaries, and technical texts. But when users co-create fictional character narratives, we see the highest similarity, indicating that they tend to ask closed-domain questions grounded in the model generation. Similarity is based on a \texttt{tf-idf} representation of texts. Vertical lines indicate the median similarity.}
    \label{fig-followup2prevgen}
\end{figure*}

\begin{figure*}
\centering
\begin{subfigure}[c]{.45\textwidth}
   \includegraphics[width=1.05\textwidth]{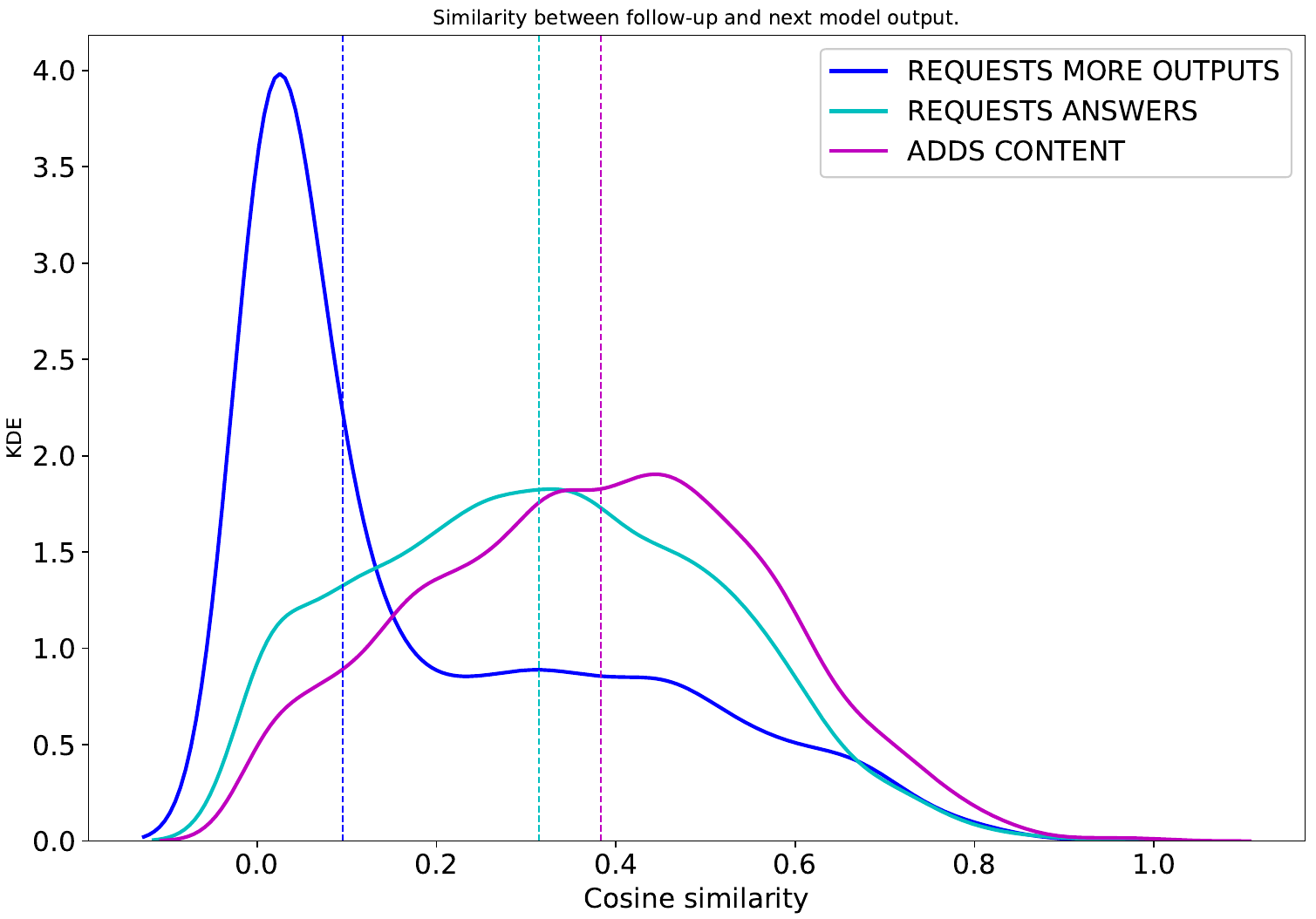}
    \caption{Similarity between follow-up and next generation in \wildchatwr for creative writing intents.}
    \label{fig-followup}
\end{subfigure}~
\begin{subfigure}[c]{.5\textwidth}
\centering
    {\includegraphics[width=0.9\textwidth]{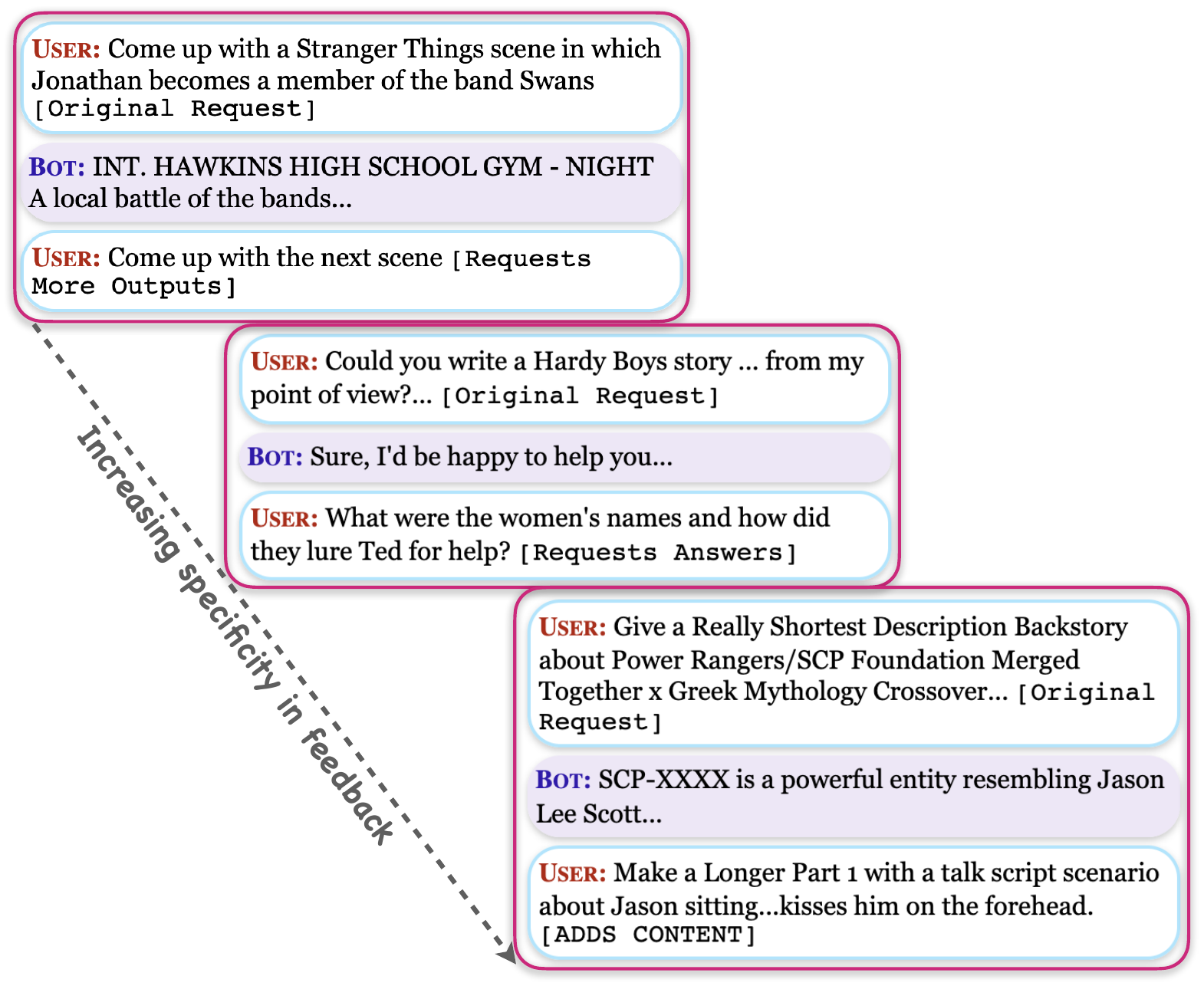}}
    \caption{Increasing specificity of feedback in user follow-ups.}
    \label{fig-morespecific-examples}
\end{subfigure}
\caption{(a) Increasing similarity between user follow-up and the next model generation across follow-up types \textsc{requests more outputs}, \textsc{requests answers}, and \textsc{adds content}. This indicates an increasing amount of specificity in user feedback across the three kinds of follow-up types. We discuss this in \S\ref{sec-discussion}. (b) Example conversations from \wildchatwr that illustrate the increasingly specific follow-up types for the writing intent \textsc{generate character}. Similarity is based on a \texttt{tf-idf} representation of texts. Vertical lines indicate the median similarity.}
\label{fig-morespecific}
\end{figure*}

We discuss how writing intents correlate with \proposedacronym{}s in \S\ref{sec-results}. We include additional results here. 
\begin{itemize}[nolistsep, noitemsep]
    \item Appendix \ref{sec-results-changestyle}: Discusses writing intents correlated with \textsc{change style}/\proposedacronym{}4 -- we include it here for a shortage of space.
    \item Appendix \ref{sec-results-revising}: Includes dataset dependent trends for \proposedacronym{}1 and 5 -- we include it here for completeness.
    \item Figures \ref{fig-supp-coeffplots}, \ref{fig-followup2prevgen}, and \ref{fig-morespecific} supplement the result (\S\ref{sec-results}) and discussion (\S\ref{sec-discussion}).
\end{itemize}

\begin{figure*}
\centering
\begin{subfigure}[c]{.45\textwidth}
   \includegraphics[trim=0.9cm 0.9cm 0cm 1.53cm, clip, width=1.05\textwidth]{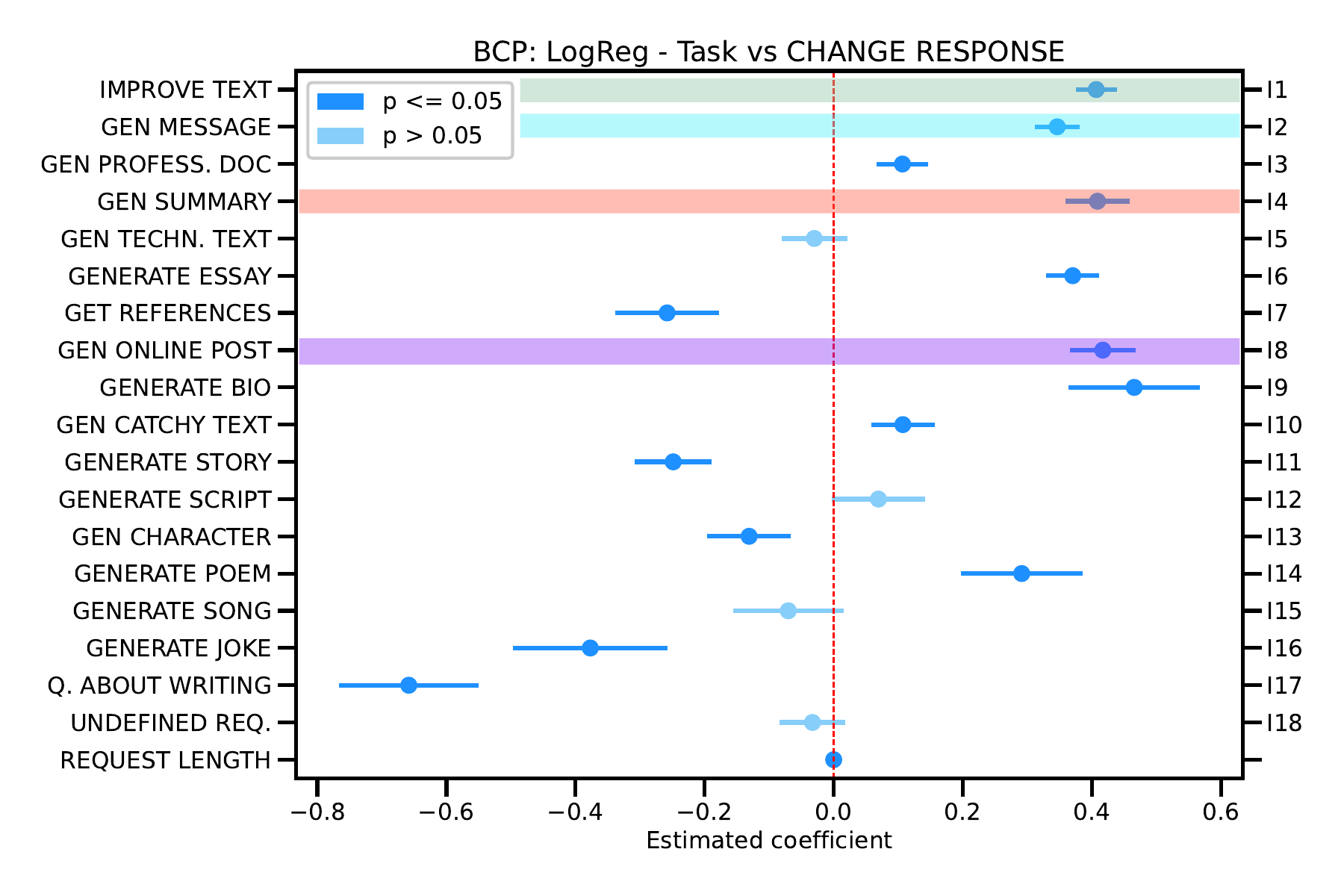}
    \caption{Logistic regression coefficients for intents vs \textsc{change style} in \bcpwr.}
    \label{fig-coefp-bcp-chsty}
\end{subfigure}~
\begin{subfigure}[c]{.5\textwidth}
\centering
    {\includegraphics[width=1.0\textwidth]{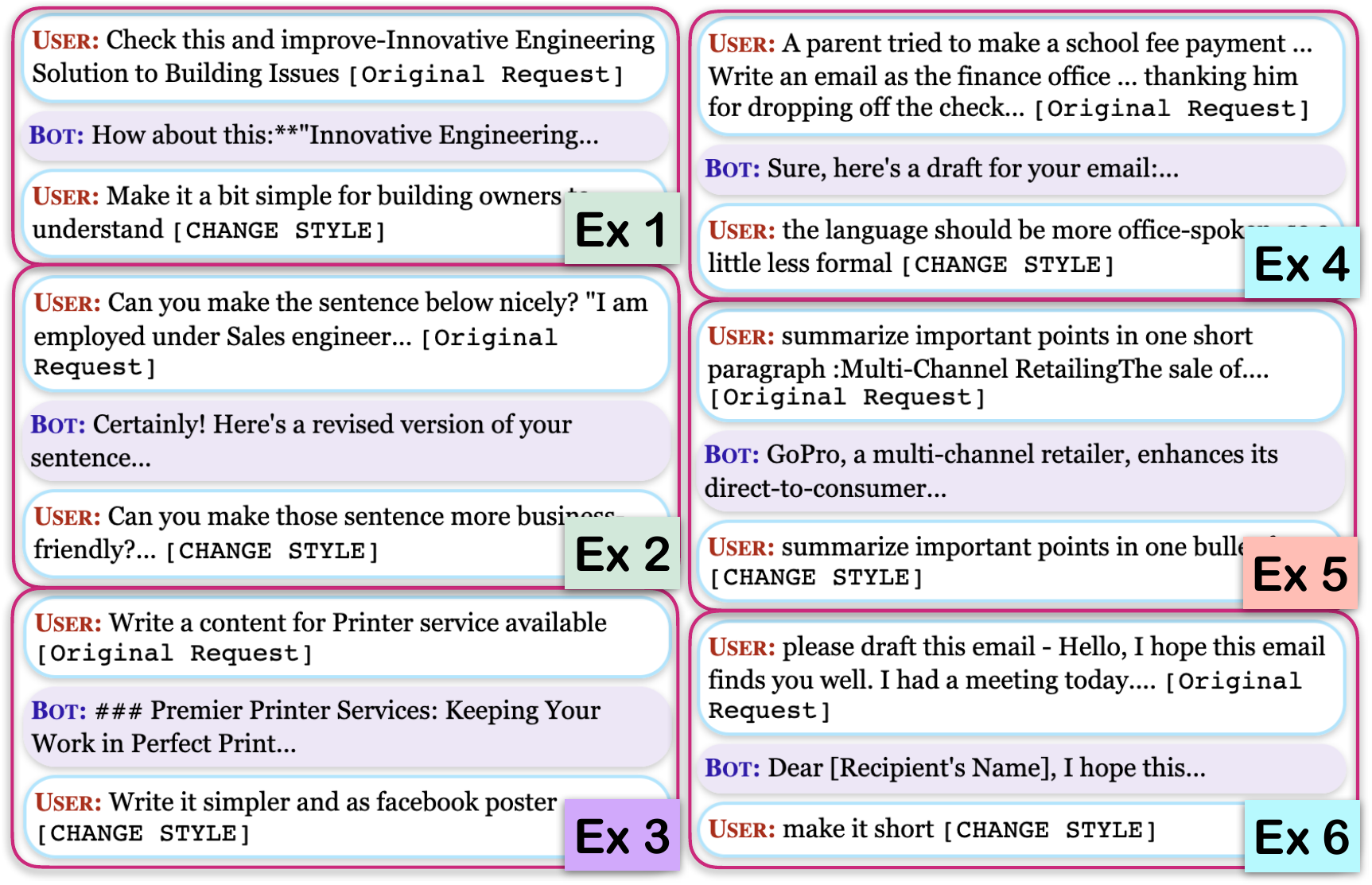}}
    \caption{Users engaged in intents I1, I2, I4, and I8.}
    \label{fig-chsty-examples}
\end{subfigure}
\caption{(a) The intents discussed in \S\ref{sec-results-changestyle} are highlighted in color. Coefficients for \wildchatwr are plotted in Figure \ref{fig-supp-coeffplots}. (b) Example conversations from the intents highlighted in (a) -- intent and example colors are matched.}
\end{figure*}
\subsection{Changing style to align generations with readers}
\label{sec-results-changestyle}
Figure \ref{fig-coefp-bcp-chsty} shows the writing intents correlated with \textsc{change style} where users modified the style of LLM generations. This \proposedacronym{} correlated with intents to improve texts across domains (I1), and generate messages, summaries, and online posts (I2, I4, I8). Analysis of sessions revealed that users frequently changed the style of generations to better match the assumed norms for a communication (Ex \greenintent{2}, \purpleintent{3}, \neonblueintent{4}) or to the likely preference of readers (Ex \greenintent{1}, \redintent{5}). Specifically, when users sought to \textsc{improve text} (I1), their follow-ups changed the style to better match the preferences of specific readers or the business context of a communication (Ex \greenintent{1}, \greenintent{2}). This remained true for communicative texts intended for groups or individuals (Ex \purpleintent{3}, \neonblueintent{4}). And when users generated summaries, they sought to change their format for readability (Ex \redintent{5}), focusing on readers. Our findings lend support to prior work on style matching in communicative settings. This work notes that style matching is crucial for engagement and community identity on online platforms \citep{ananthasubramaniam2023exploring, tran2016characterizing}, and an important element of effective communication \citep{pickering2013integrated}. Our work extends style matching to the case of user-LLM co-creation. Inferring and aligning LLMs to aid co-creation for specific audiences remains under-explored -- this may be seen as a form of reader personalization. Finally, across intents we also noted \textsc{change style} to request more concise outputs (Ex \neonblueintent{6}) -- this may be attributed to RLHF aligned models' generating lengthy outputs \citep{singhal2024a}.

\textit{\textbf{Implications:}} 
\begin{itemize*}
    \item Infer intended readers and their preferred style, enabling LLMs to customize the generation style to them.
\end{itemize*}

\subsection{Dataset dependent trends in revising and elaborating on requests}
\label{sec-results-revising}
\begin{figure*}
\centering
\begin{subfigure}[c]{.45\textwidth}
   \includegraphics[trim=0.9cm 0.9cm 0cm 1.53cm, clip, width=1.05\textwidth]{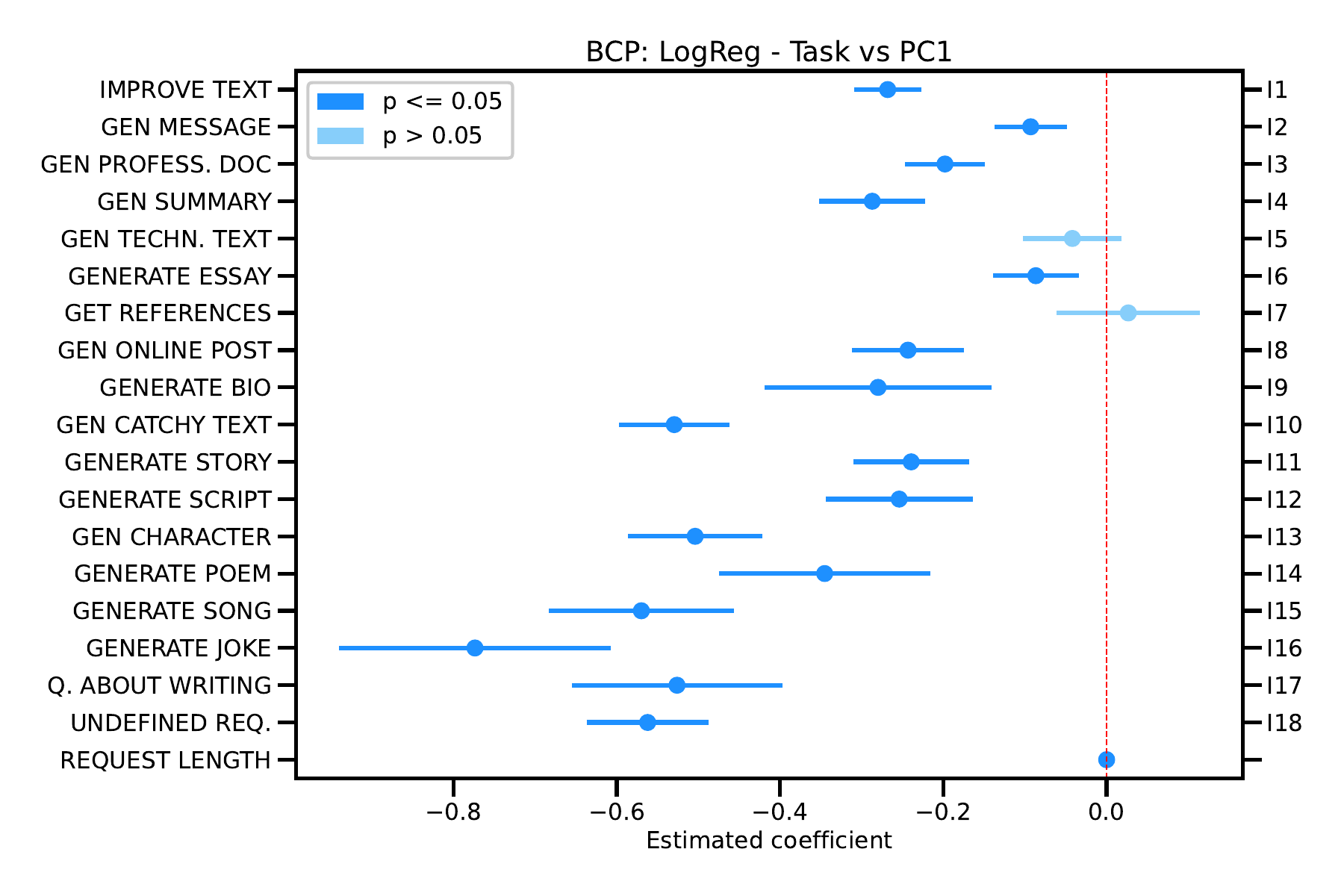}
    \caption{Writing intents vs \proposedacronym{}1 in \bcpwr.}
    \label{fig-coefp-bcp-pc1}
\end{subfigure}~
\begin{subfigure}[c]{.45\textwidth}
\includegraphics[trim=0.9cm 0.9cm 0cm 1.53cm, clip, width=1.05\textwidth]{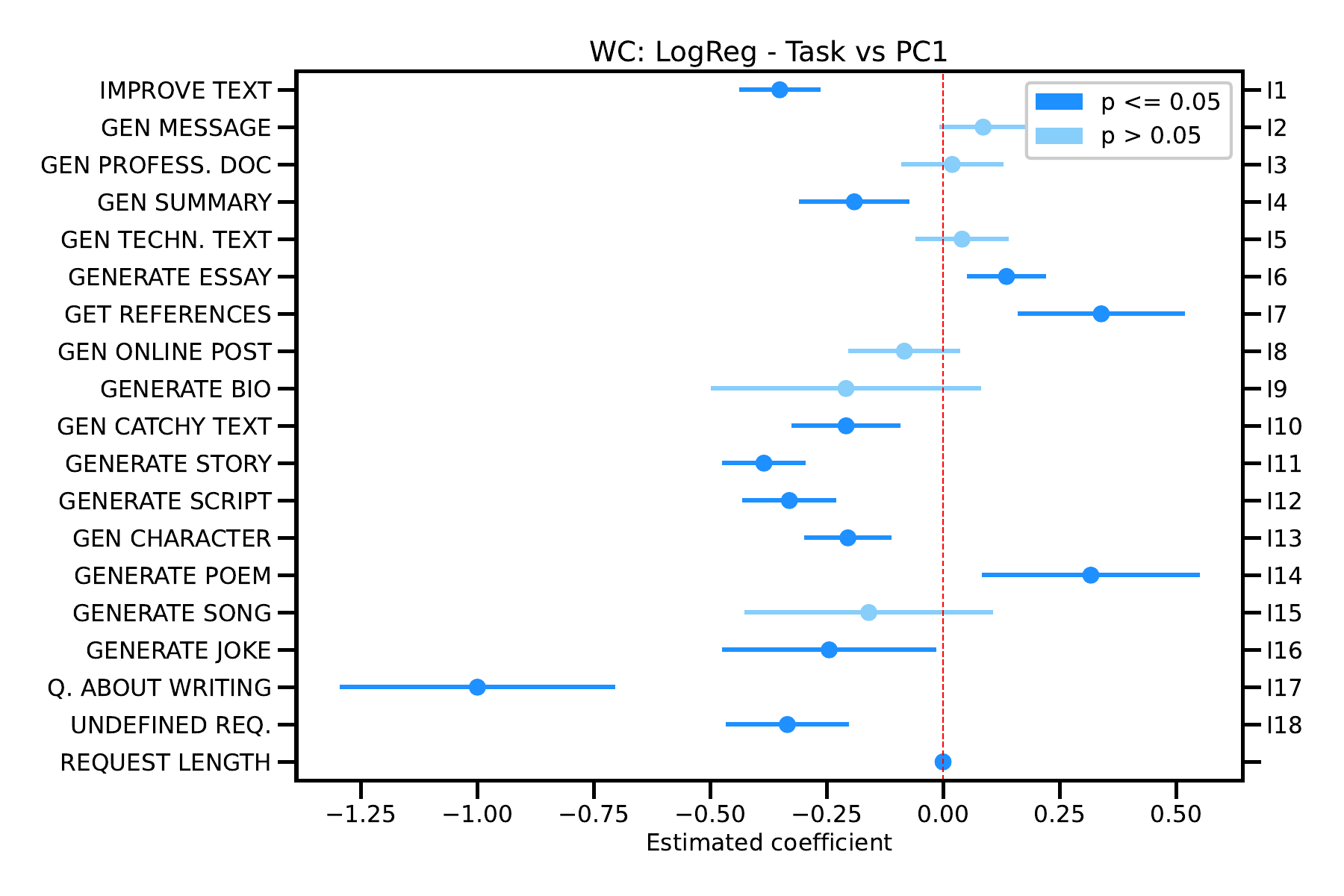}
    \caption{Writing intents vs \proposedacronym{}1 in \wildchatwr.}
    \label{fig-coefp-wc-pc1}
\end{subfigure}

\caption{Coefficient plots for writing intents vs \proposedacronym{}1 (revising requests).}
\label{fig-supp-coeffplots-pc1}
\end{figure*}

\begin{figure*}
\centering
\begin{subfigure}[c]{.45\textwidth}
   \includegraphics[trim=0.9cm 0.9cm 0cm 1.53cm, clip, width=1.05\textwidth]{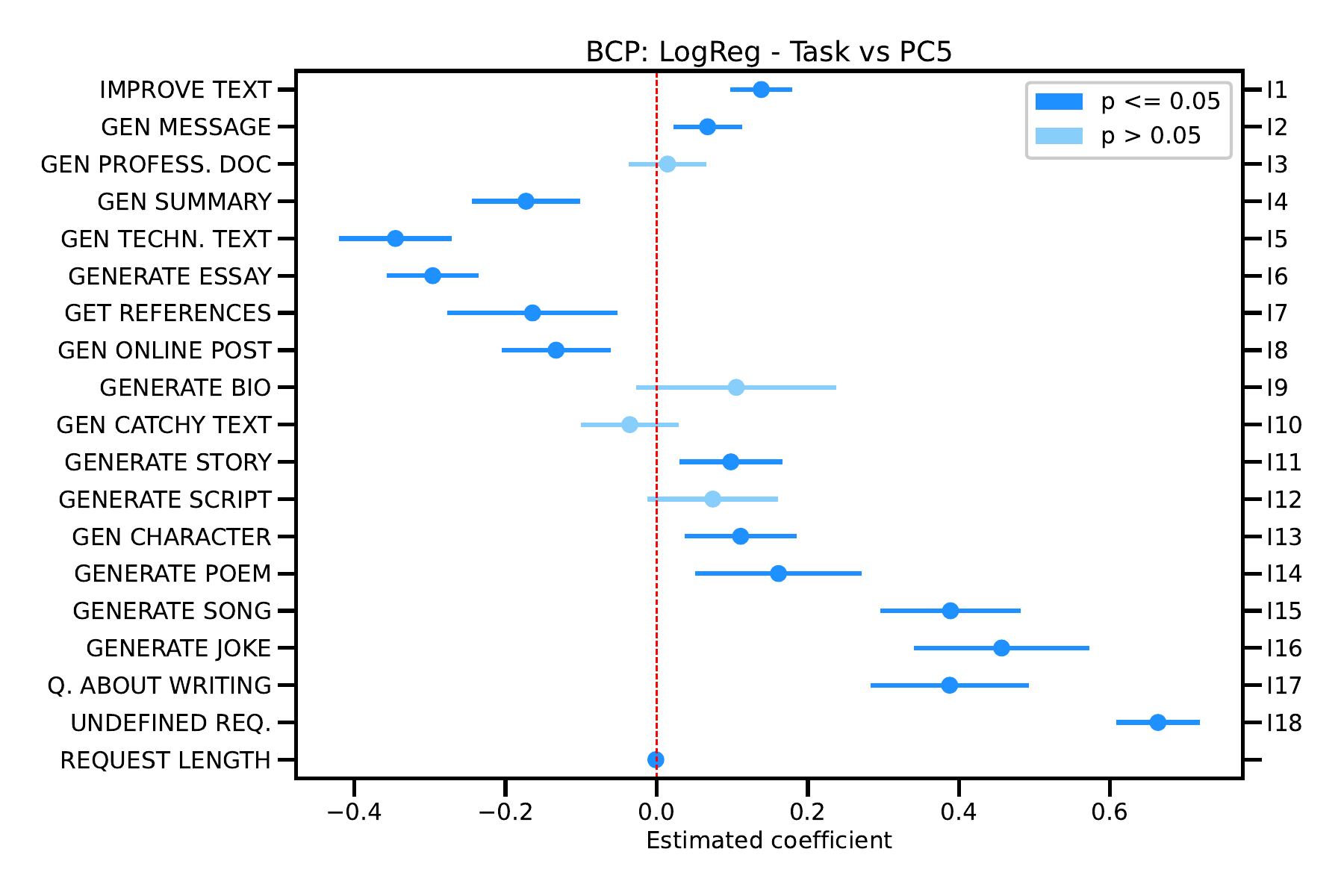}
    \caption{Writing intents vs \proposedacronym{}5 in \bcpwr.}
    \label{fig-coefp-bcp-pc6}
\end{subfigure}~
\begin{subfigure}[c]{.45\textwidth}
\includegraphics[trim=0.9cm 0.9cm 0cm 1.53cm, clip, width=1.05\textwidth]{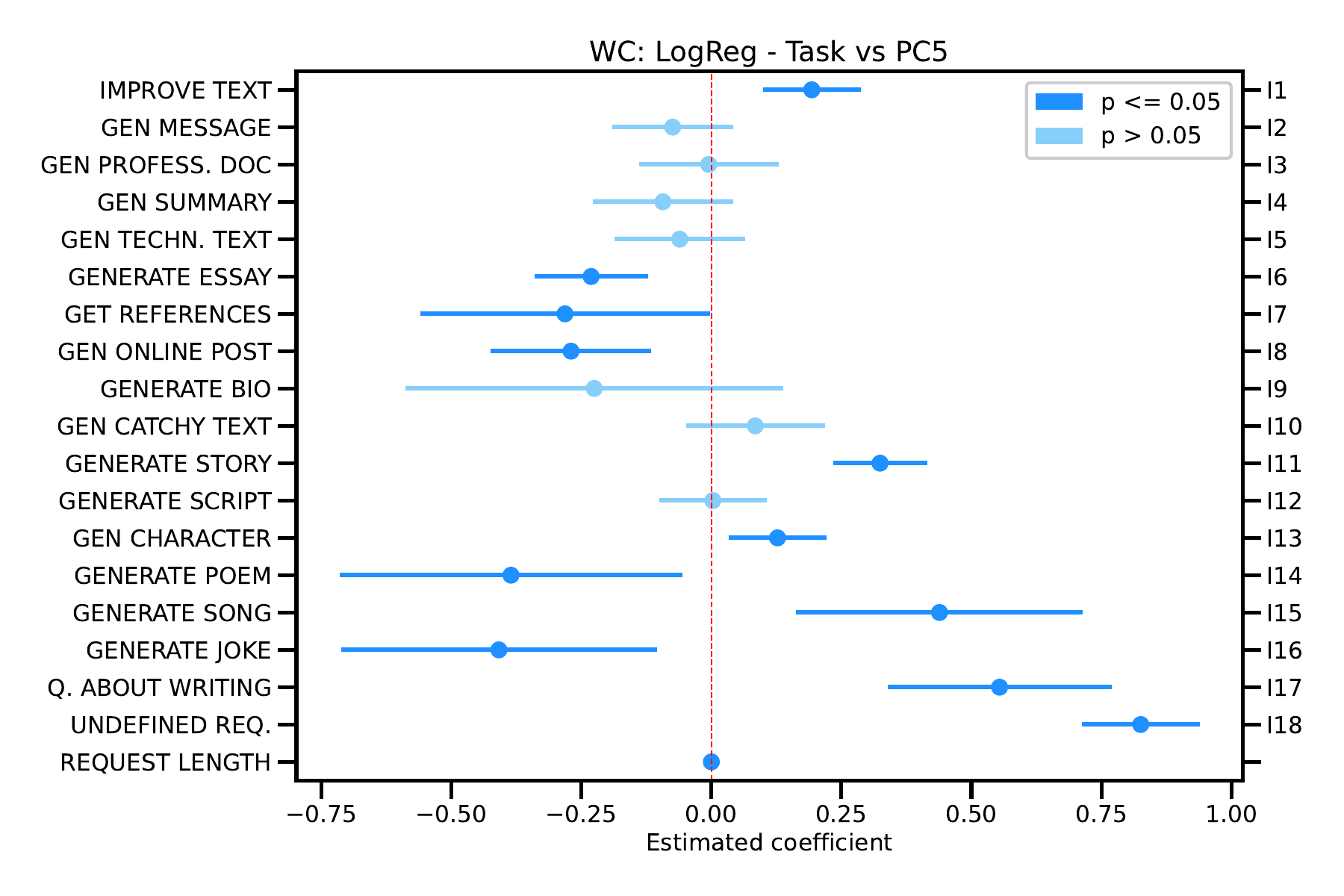}
    \caption{Writing intents vs \proposedacronym{}5 in \wildchatwr.}
    \label{fig-coefp-wc-pc6}
\end{subfigure}

\caption{Coefficient plots for writing intents vs \proposedacronym{}5 (Elaborating on requests).}
\label{fig-supp-coeffplots-pc5}
\end{figure*}

Figure \ref{fig-supp-coeffplots-pc1} and \ref{fig-supp-coeffplots-pc5} show writing intents correlated with \proposedacronym{}1 and \proposedacronym{}5. We avoid discussing them in detail and include them for completeness. Respectively, they represent users revising original requests and elaborating on requests in follow-ups. For both \proposedacronym{}s, we don't see positively correlated and statistically significant intents across \bcpwr and \wildchatwr. We hypothesize that the dataset-dependent correlation with writing intents in \proposedacronym{}1 may be because users provide many different forms of feedback in a revised prompt \cite{chen2021qreformulation}, making correlation with specific writing intents less likely. Further, we also hypothesize that both \proposedacronym{}1 and 5 may be closely tied to user satisfaction, which is likely to be system dependent. In this regard, recent work \cite{sarkar2025promptrevision} leverages the feedback contained in revised requests to improve LLM performance. Future work may attempt to identify more distinguished \proposedacronym{}s through a finer-grained analysis of sessions with revised or elaborated requests.

\clearpage
\begin{figure*}
\centering
\begin{subfigure}[c]{.45\textwidth}
   \includegraphics[trim=0.9cm 0.9cm 0cm 1.53cm, clip, width=1.05\textwidth]{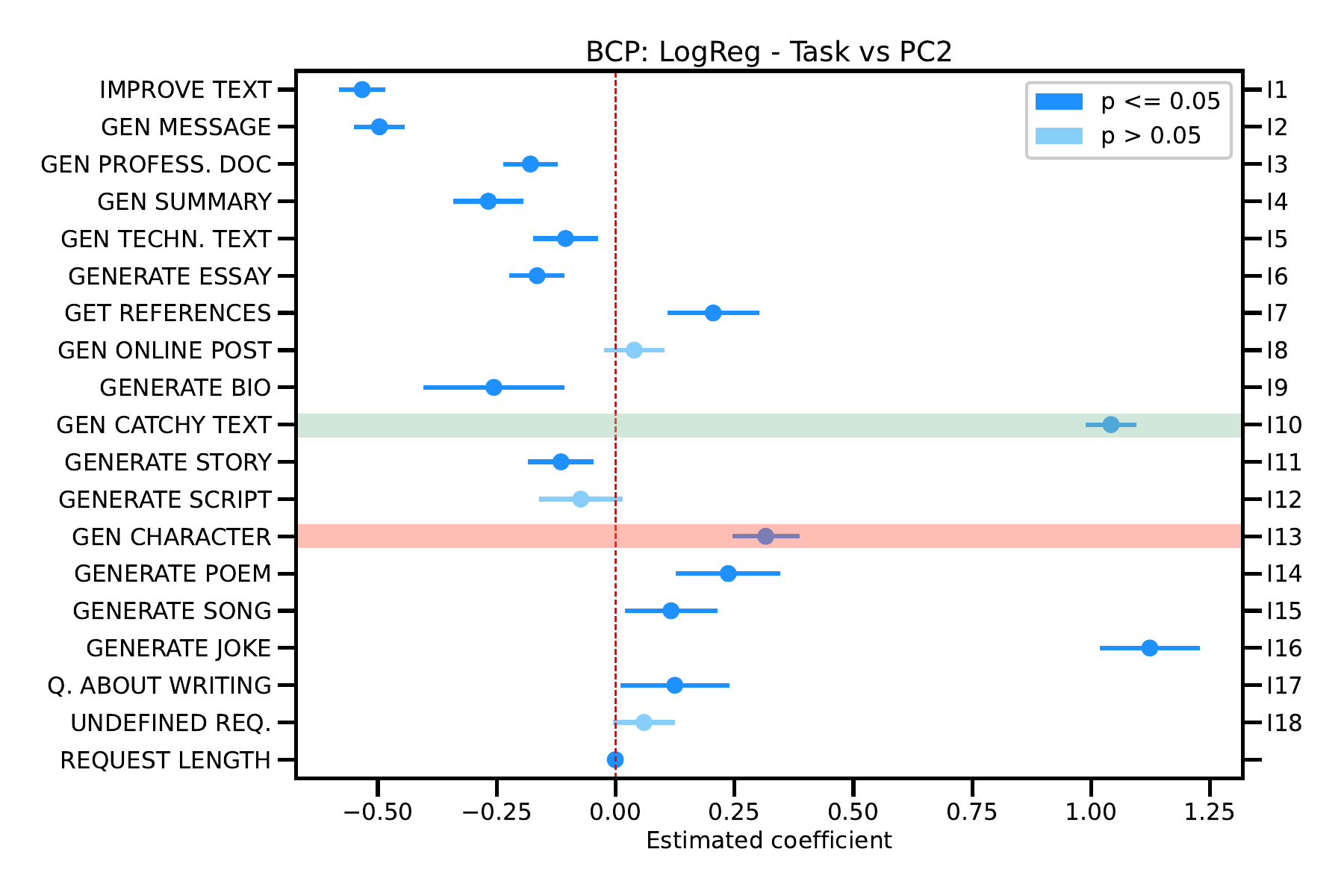}
    \caption{Writing intents vs \proposedacronym{}2 in \bcpwr.}
    \label{fig-coefp-bcp-pc2}
\end{subfigure}~
\begin{subfigure}[c]{.45\textwidth}
\includegraphics[trim=0.9cm 0.9cm 0cm 1.53cm, clip, width=1.05\textwidth]{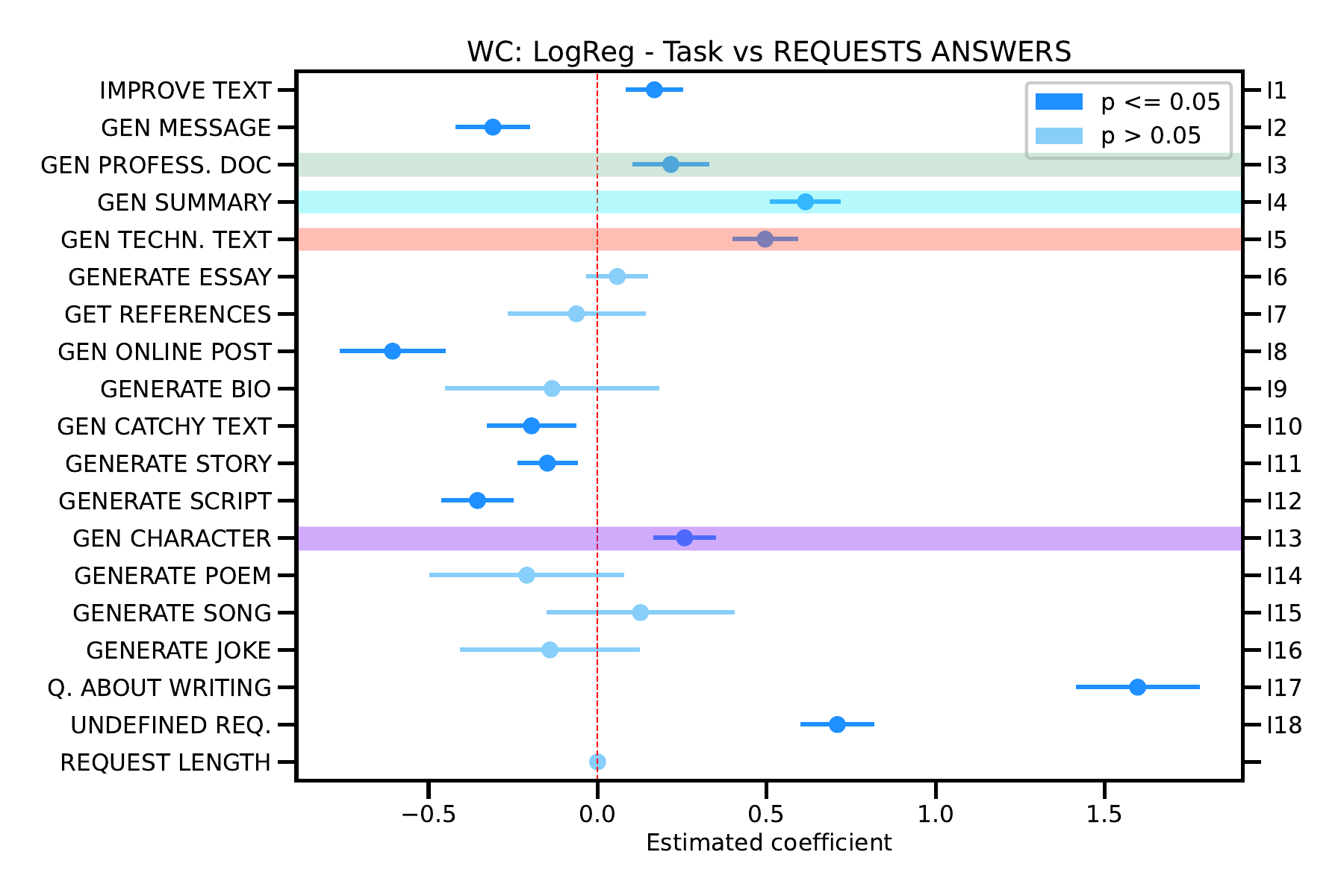}
    \caption{Writing intents vs \proposedacronym{}3 in \wildchatwr.}
    \label{fig-coefp-wc-pc3}
\end{subfigure}\\[1cm]

\begin{subfigure}[c]{.45\textwidth}
   \includegraphics[trim=0.9cm 0.9cm 0cm 1.53cm, clip, width=1.05\textwidth]{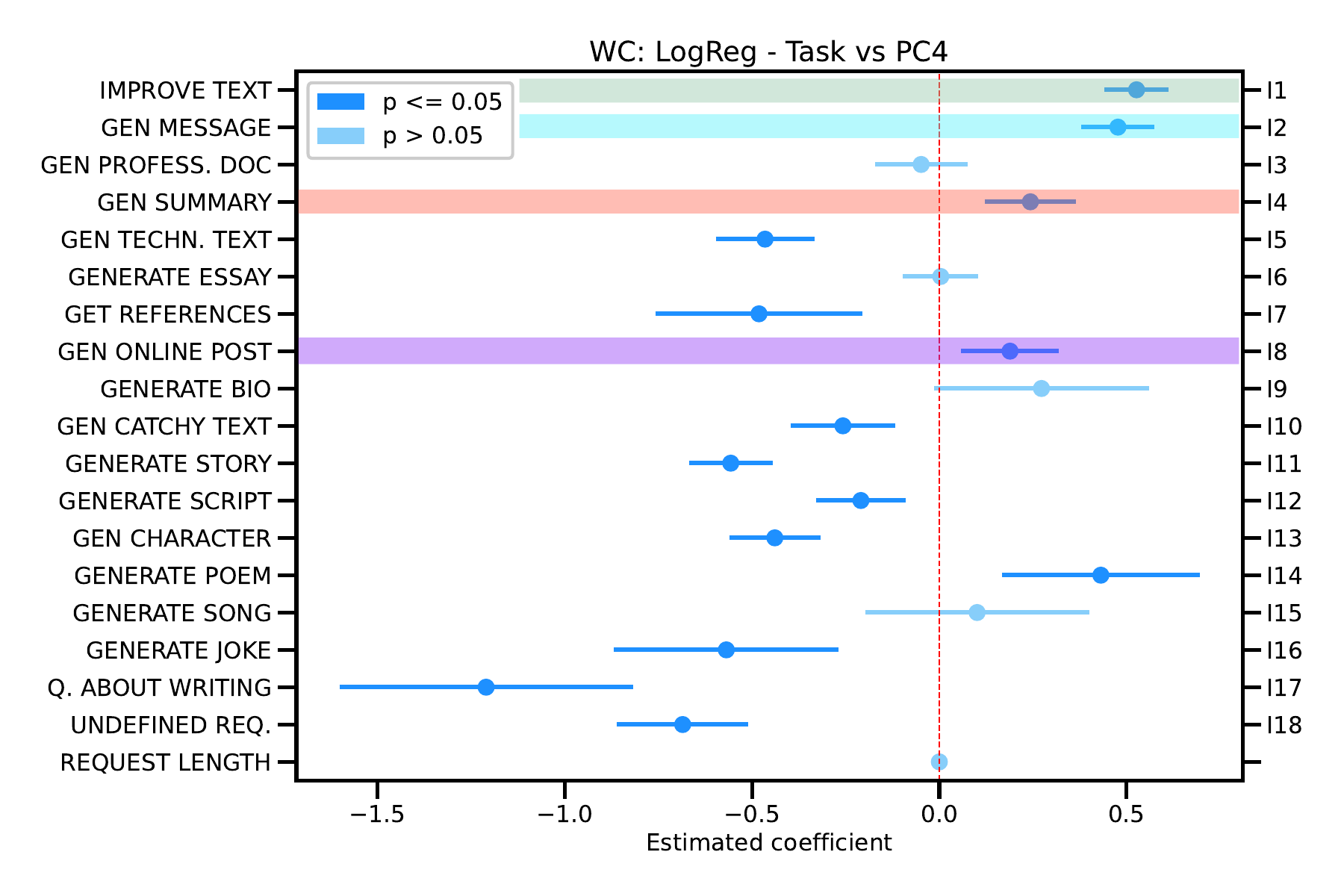}
    \caption{Writing intents vs \proposedacronym{}4 in \wildchatwr.}
    \label{fig-coefp-wc-pc4}
\end{subfigure}~
\begin{subfigure}[c]{.45\textwidth}
\includegraphics[trim=0.9cm 0.9cm 0cm 1.53cm, clip, width=1.05\textwidth]{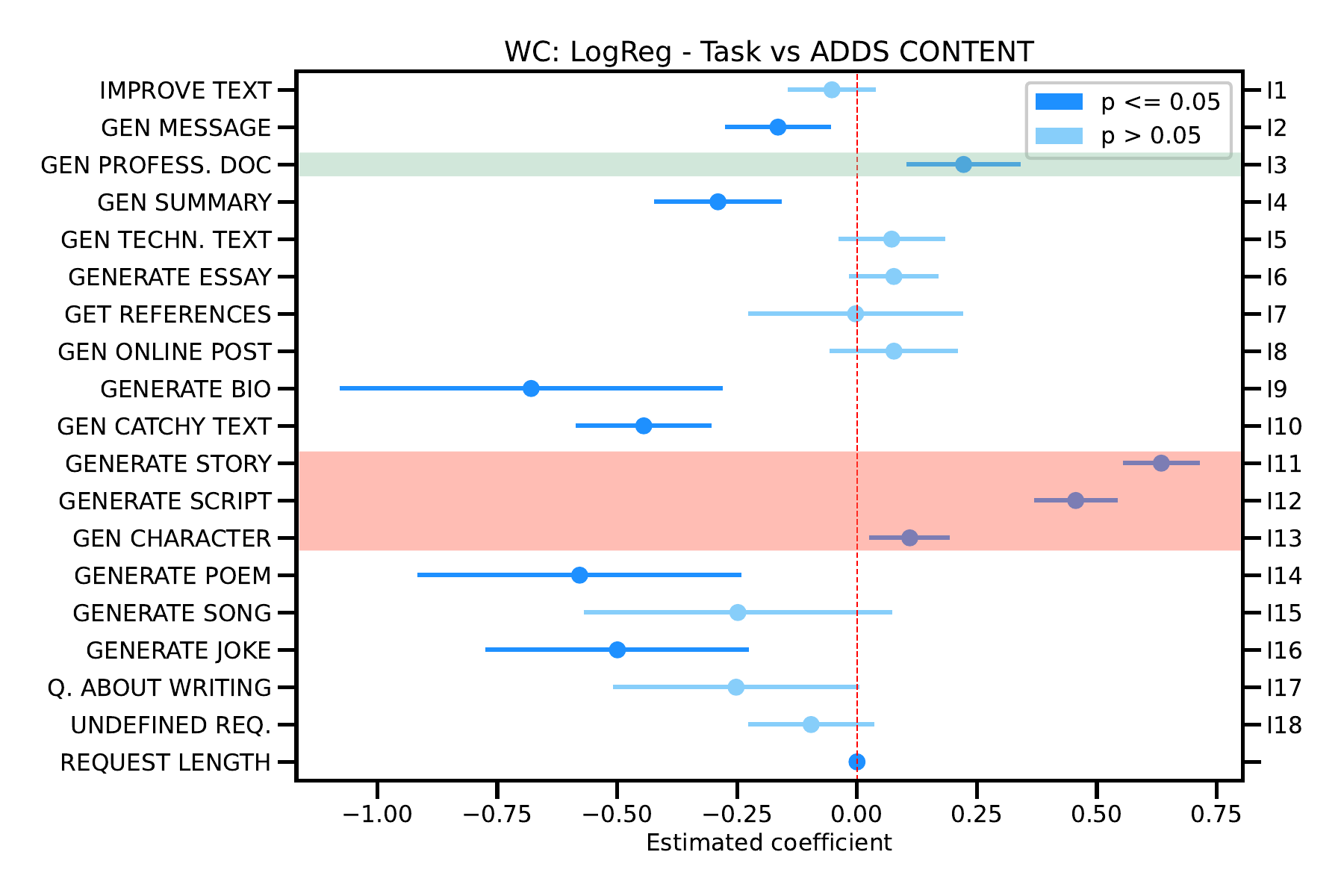}
    \caption{Writing intents vs \textsc{adds content} in \wildchatwr.}
    \label{fig-coefp-wc-addcon}
\end{subfigure}

\caption{Supplementary coefficient plots for Figure \ref{fig-coefp-wc-pc2}, Figure \ref{fig-coefp-bcp-pc3}, Figure \ref{fig-coefp-bcp-chsty}, and Figure \ref{fig-coefp-bcp-addcon}. Colored intents are discussed in \S\ref{sec-results} and match the colors of their corresponding coefficient plots and examples in \S\ref{sec-results}.}
\label{fig-supp-coeffplots}
\end{figure*}
\clearpage
\clearpage
\begin{lstfloat*}[t]
\begin{lstlisting}[breakautoindent=false, breaklines=true, breakindent=0pt]
# Task Instructions
You an an experienced linguist who helps analyze conversations. Given a CONVERSATION between a user and an assistant, classify the CONVERSATION into one or more of the following labels.
INFORMATION SEARCH: The user is asking a question about a specific document or could have searched for this information on the internet. +$\leftarrow$ \textcolor{PineGreen}{Search Information}+
GENERATING COMMUNICATIVE TEXT: The user is trying to generate text for communication with an individual, a group, or an online platform. +$\leftarrow$ \textcolor{PineGreen}{Gen Communicative Text}+
SOFTWARE DEVELOPMENT: The user is seeking assistance in software development. Often phrased as a how to question with accompanying code. +$\leftarrow$ \textcolor{PineGreen}{Develop Software}+
HOMEWORK PROBLEM: The user is posing a question or problem which is likely to be from an exam or homework. +$\leftarrow$ \textcolor{PineGreen}{Solve Problem}+
GENERATING CREATIVE TEXT: The user is seeking to generate creative stories, characters, titles, slogans and other creative texts. +$\leftarrow$ \textcolor{PineGreen}{Generate Creative Text}+
GENERATING TECHNICAL TEXT: The user is working with the assistant to generate technical or research texts. +$\leftarrow$ \textcolor{PineGreen}{Generate Technical Text}+
SUMMARIZATION: The user seeks to summarize a text that they provide. +$\leftarrow$ \textcolor{PineGreen}{Summarize Text}+
TRANSLATION: The user is seeking to translate a text or understand text in a language that isnt English. +$\leftarrow$ \textcolor{PineGreen}{Translate Text}+
IMAGE GENERATION: The user asks to generate an image or other forms of visual art.  +$\nwarrow$ \textcolor{PineGreen}{Generate Image}+
EXPLICIT CONTENT: The user or bot text contains explicit or pornographic content.  +$\nwarrow$ \textcolor{PineGreen}{Generate Explicit Text}+
UNDEFINED: A request for which none of the defined labels are applicable or there isn't an explicit request by the USER. +$\leftarrow$ \textcolor{PineGreen}{Undefined Request}+

For your response use the following instructions:
1. Output one of more of the correct labels for the CONVERSATION
2. Output the labels in decreasing order of their relevance to the CONVERSATION

CONVERSATION:
+\textbf{\{\{conversation\_text\}\}}+
Output a explanation and then one or more of the labels for the CONVERSATION.
\end{lstlisting}
\caption{The prompt for GPT-4o based multi-label Task Classifier (\coarsetaskclf) used to obtain the writing task sessions from \bcpall and \wildchatall. This is discussed in Section \ref{sec-analysis-setup} and Appendix \ref{sec-analysis-setup-supp}. The \textcolor{PineGreen}{text in green} maps the labels used in the prompt to the labels used in Figure \ref{tab-exploratory-session-stats} and the rest of the paper. \textbf{Bold text} is replaced with raw conversation data.}
\label{fig-coarsetaskclf-prompt}
\end{lstfloat*}
\clearpage
\begin{lstfloat*}[t]
\begin{lstlisting}[breakautoindent=false, breaklines=true, breakindent=0pt]
# Task Instructions
You an an experienced linguist who helps analyze conversations. Given a conversation between a USER and an ASSISTANT, label the USER UTTERENCES with the below labels.
Use the following labels:
NEW REQUEST: The user is making a new request to the ASSISTANT for a new topic and for the first time. +$\leftarrow$ \textcolor{PineGreen}{ORIGINAL REQUEST}+
RESTATES REQUEST: The user is restating the NEW REQUEST with a modification.
ELABORATES REQUEST: The user is elaborating on the NEW REQUEST without restating 
the request.
COURTESY RESPONSE: The user is responding as a courtesy to the ASSISTANT (e.g. "yes please", "go ahead" etc) or is exchanging pleasantries with the ASSISTANT.
CHANGE RESPONSE: The user is requesting the ASSISTANT to change the stylistic elements (e.g. length, tone, formality etc) of its previous response. +$\leftarrow$ \textcolor{PineGreen}{CHANGE STYLE}+
ADDS CONTENT: The user is requesting new content to be added to the ASSISTANTs response.
REMOVES CONTENT: The user is requesting specific content to be removed from the ASSISTANTs response.
REQUESTS ANSWERS: The user is requesting new or additional information related to the ASSISTANTs response.
REQUESTS MORE OUTPUTS: The user is requesting additional output from the ASSISTANT.
RESPONDS POSITIVELY: The user is responding positively to the ASSISTANTs response without providing additional detail.
RESPONDS NEGATIVELY: The user is responding negatively to the ASSISTANTs response without providing additional detail.
UNDEFINED RESPONSE: None of the defined labels describe the users response.

Use these rules when you respond:
1. Output one best label for every USER UTTERENCE.
2. Output the labels in the order of the USER UTTERENCES.

Here is the conversation:
+\textbf{\{\{conversation\_text\}\}}+
Output a label for each USER UTTERENCE.
\end{lstlisting}
\caption{The prompt for GPT-4o used to label user utterances as original requests and or one among eleven follow-up types in \bcpwr and \wildchatwr. This is discussed in Section \ref{sec-method} and Appendix \ref{sec-method-identifypref-supp}. The \textcolor{PineGreen}{text in green} maps the labels used in the prompt to the labels used in Table \ref{tab-follow-up-descr} and the rest of the paper if they differ. \textbf{Bold text} is replaced with raw conversation data.}
\label{fig-fgfollowup-prompt}
\end{lstfloat*}
\clearpage
\begin{table*}[t]
\scalebox{0.82}{
\begin{tabular}{>{\raggedleft\arraybackslash}p{1.7in}p{1.7in}p{3.7in}}
\toprule
Follow-up Type & Description & Examples (\orexample{original request} and \fupexample{follow-up})\\
\midrule

\textsc{restates request} & Reformulates their request & \orexample{``Msitu Africa is requesting for funding through crowdfunding websites. Write for me a small compelling write up requesting for funds.''} $\rightarrow$ \fupexample{``Fundraiser Story. Msitu Africa is requesting for funding through crowdfunding websites. Write for me a small compelling write up requesting for funds to help us actualize our initiative.''}\\
& & \orexample{``wrire a detail script for animation video about Pan Am Flight 103''} $\rightarrow$ \fupexample{``write a detail script for animation video about Pan Am Flight 103, video will be 20 minutes in length''}\\
\rowcolor{gray!20}
\textsc{elaborates request} & Expands on their request; often after being asked by the LLM & \orexample{Can you help me with writing the discussion for my article?} $\rightarrow$ \fupexample{``The article objectives are to explore and analyze the various impacts of D-penicillamine (DPCA) treatment...''}\\
\rowcolor{gray!20}
& & \orexample{``i want to provide you with 4 separate inputs as follows story1 chapter1, story1 chapter7, story2 chapter7, i would like you to write story2...''} $\rightarrow$ \fupexample{``Chapter 1: Shadows of Rebellion...''}\\

\textsc{requests answers} & Question related to the output & \orexample{``My potential client has just returned from a trip last week. How should I greet him in an email''} $\rightarrow$ \fupexample{``What should I write in the subject line in my email''}\\
& & \orexample{``I want you to write a story for me. There are two girl who hate each other. They are going to wrestling in front of their friends...''} $\rightarrow$ \fupexample{``can you describe me last pin position ?''}\\

\rowcolor{gray!20}
\textsc{requests more outputs} & Asks for additional output & \orexample{``Write 10 funny and short answers to this comment: ...''} $\rightarrow$ \fupexample{``Write more''}\\
\rowcolor{gray!20}
& & \orexample{``you are a novelist. settings - late medieval england like kingdom. kingdoms A and B. ... list 25 approaches A can benefit from''} $\rightarrow$ \fupexample{``Give 25 more''}\\

\textsc{change style} & Changes style of output & \orexample{``rewrite this Hotel A is an excellent choice for tourists seeking a memorable and enjoyable stay...''} $\rightarrow$ \fupexample{``rewrite it in 240 words''}\\
& & \orexample{`` write an email to change the name of the HR manager, which due to mistake on my part was spelled incorrectly''} $\rightarrow$ \fupexample{``write something instead of Dear''}\\

\rowcolor{gray!20}
\textsc{adds content} & Adds content to output & \orexample{``write in bullet points my responsibilities as an intern during a neurology rotation''} $\rightarrow$ \fupexample{``mention discussions''}\\
\rowcolor{gray!20}
& & \orexample{``script about cincinnati blowing a 21 point lead vs johnny manziel''} $\rightarrow$ \fupexample{``the next week seattle blew out cincinnaty by more than 30''}\\

\textsc{removes content} & Remove content from output & \orexample{``Make this into a episode and give it More life''} $\rightarrow$ \fupexample{``Remove panel 1-4 and re number the rest''}\\
& & \orexample{``Today is my senior patrol election and I want to win. I need to provide a 2-3 minute speech...''} $\rightarrow$ \fupexample{``Remove the details about having less experience and emphasize how I want to use my creativity...''}\\

\rowcolor{gray!20}
\textsc{courtesy response} & A courtesy or pleasantry & \orexample{``write a short 100 word Youtube channel description for a channel...''} $\rightarrow$ \fupexample{``thank you''}\\
\rowcolor{gray!20}
& & \orexample{``Can you write a 300 pages English novel if I give you the plot''} $\rightarrow$ \fupexample{``Ok so can we start''}\\

\textsc{responds positively} & Explicitly pleased with output & \orexample{``In my book there is a central plot with the main characters who influence global events...''} $\rightarrow$ \fupexample{``excellent, keep going''}\\
& & \orexample{``i want to create 4 different kinds of postings. one should be Good morning''} $\rightarrow$ \fupexample{``i like them plase continue with 10 for each type''}\\

\rowcolor{gray!20}
\textsc{responds negatively} & Explicitly unhappy with output & \orexample{``Could you come up with some two syllables long feminine robot names? They need to be based...''} $\rightarrow$ \fupexample{``Most of those where three syllables long''}\\
\rowcolor{gray!20}
& & \orexample{``Write a brief man to man heartfelt reply to this comment...''}$\rightarrow$ \fupexample{``Do you not understand brief? No yapping''}\\
\bottomrule
\end{tabular}
}
\caption{Example utterances for each follow-up type selected from \wildchatwr. We include the \orexample{original requests} to better illustrate the \fupexample{follow-up} utterance by the user. We omit the LLM response for space.} 
\label{tab-follow-up-examples}
\end{table*}
\clearpage
\begin{lstfloat*}[t]
\begin{lstlisting}[breakautoindent=false, breaklines=true, breakindent=0pt]
# Task Instructions
You an an experienced linguist who helps analyze conversations. 
Given a USER REQUEST to an assistant, classify the USER REQUEST into one or more of the following labels:
GENERATE SONG: Request to write or re-write a song.
GENERATE JOKE: Request to write or re-write a joke.
GENERATE POEM: Request to write or re-write a poem, haiku or a similar other verse.
GENERATE STORY: Request to write or re-write a story.
GENERATE SCRIPT: Request to write or re-write a script for a video, drama, movie or similar other media.
GENERATE CHARACTER: Request to generate a fictional character and their story.
GENERATE CASUAL BIO: Request to write or re-write a bio for an online or mobile app profile. +$\leftarrow$ \textcolor{PineGreen}{GENERATE BIO}+
GENERATE PROFESSIONAL DOCUMENT: Request to write or re-write a resume, recommendation letter, or other professional document.
GENERATE MESSAGE: Request to write or re-write a interpersonal message, cover letter, email, letter or other interpersonal communication.
GENERATE ONLINE POST: Request to write or re-write a text for an online platform like a social media post, review etc.
GENERATE TITLE: Request to write a title, slogan, or other eye catching text. 
 +$\nwarrow$ \textcolor{PineGreen}{GENERATE CATCHY TEXT}+
GENERATE SUMMARY: Request to summarize a text provided by the user.
GENERATE ESSAY: Request to write an essay, blog, article or other long text on any topic.
GENERATE TECHNICAL TEXT: Request to write or re-write a scientific or technical text
QUESTION ABOUT WRITING: A question about literary works, writing or publishing.
IMPROVE TEXT: Request to improve grammar, style, tone or other aspect in the provided text.
GET REFERENCES: Request to add, format or generate references.
UNDEFINED REQUEST: A request for which none of the labels are applicable, or there isn't an explicit request, or a request that a writing assistant cannot fulfill.

For your response use the following instructions:
1. Output one of more of the correct labels for the USER REQUEST.
2. Output the labels in decreasing order of their relevance to the USER REQUEST.

USER REQUEST: +\textbf{\{\{user\_request\}\}}+
Output a explanation and then one or more of the labels for the USER REQUEST.
\end{lstlisting}
\caption{The prompt for GPT-4o used to label original requests with finer-grained writing intents \bcpwr and \wildchatwr. This is discussed in Section \ref{sec-method} and Appendix \ref{sec-method-correlatepref-supp}. The \textcolor{PineGreen}{text in green} maps the labels used in the prompt to the labels used in Table \ref{tab-fgtype-descr} and the rest of the paper if they differ. Bold text is replaced with raw conversation data.}
\label{fig-fgintents-prompt}
\end{lstfloat*}
\clearpage
\begin{table*}[t]
\scalebox{0.82}{
\begin{tabular}{>{\raggedleft\arraybackslash}p{2.1in}p{4.8in}}
\toprule
Writing intent type & Example original requests\\
\midrule
\textsc{improve text} &  \orexample{``say better and more professional: Beckey, really appreciate your advocacy during''}\\
& \orexample{``Can you improve and make a variation of this sentence?: Is to be expected of any aspect...''}\\

\rowcolor{gray!20}
\textsc{generate message} & \orexample{``Hi, May I have your updated delivery schedule for face plate ... How to reply to customer and we have stock 250pcs can arrange delivery on 16 jan 2024.''}\\
\rowcolor{gray!20}
& \orexample{``write a one sentence valentine card message for a girl who bought a new pontiac...''}\\

\textsc{generate professional} & \orexample{``create a school policy regarding parents using a school account ...''}\\
\textsc{document} & \orexample{``I am writing an annual performance appraisal for my direct report who is a sales engineer...''}\\

\rowcolor{gray!20}
\textsc{generate summary} & \orexample{``give introduction to my chapter 5 (conclusion and future research) based on this text: Discussion: This study utilized...''}\\
\rowcolor{gray!20}
& \orexample{``Make a summary of this including tools and what they are used for: A version control system...''}\\

\textsc{generate technical} & \orexample{``write me a report about miss feeding amc ... in 300 words''}\\
\textsc{text} & \orexample{``I am writing a project proposal for seed funding for an engineering project ... for my project “Waveguard” for about 1000 words using ... if there are extra information needed, please let me know.''}\\

\rowcolor{gray!20}
\textsc{generate essay} & \orexample{``Write me a 500 word essay about zodiac serial killer''}\\
\rowcolor{gray!20}
& \orexample{``Write a blog post where Pop in France has picked up several new shows...''}\\

\textsc{get references} & \orexample{``Please explain, with examples, and in detail, why Western cultures tend to value a sun tan ... Please cite sources, APA style.''}\\
& \orexample{``as a postgraduate student I want a literature review to my scientific paper ... do the references.''}\\

\rowcolor{gray!20}
\textsc{generate online post} & \orexample{``Heart Shaped Pendant Necklaces ... This is the title for a jewelry I am selling on etsy. Write a short and descriptive description that is intended for a younger teen audience...''}\\
\rowcolor{gray!20}
& \orexample{``Write 10 short tweets about today’s rainy morning at the beach from one of the most windy places''}\\

\textsc{generate bio} & \orexample{``Write me a bio for me 450 charters of someone who's loves to read books''}\\
& \orexample{``i want to write a short biography about myself saying that i am an audio engineer that graduated from...''}\\

\rowcolor{gray!20}
\textsc{generate catchy text} & \orexample{``generate 5 new original desalination company names''}\\
\rowcolor{gray!20}
& \orexample{``suggest me 10 thumbnail texts for this youtube video Young Boy Transfers...''}\\

\textsc{generate story} & \orexample{``Can you write a hypothetical what if alternate history scenario, what if Russia never sold Alaska to the US...''}\\
& \orexample{``Once upon a time, in the heart of a vast national park, there lived a dedicated and diligent park worker ... throw a party to celebrate it's recovery''}\\

\rowcolor{gray!20}
\textsc{generate script} & \orexample{``Write a funny dialogue where /co/ anon asks Valerie to go to a cottage...''}\\
\rowcolor{gray!20}
& \orexample{``I have a youtube short Idea I need you to write a script. It is legendary 1v1s...''}\\

\textsc{generate character} & \orexample{``Make a Really Shortest Description Backstory about SpongeBob SquarePants...''}\\
& \orexample{``Describe rogue gas giant named Silicodum, it has frozen dense hazes of silicate...''}\\

\rowcolor{gray!20}
\textsc{generate poem} & \orexample{``Write me a funny poem about my brother malachi...''}\\
\rowcolor{gray!20}
& \orexample{``Turn this into an unrhyming poem...''}\\

\textsc{generate song} & \orexample{``make a rap song to pony by geniune about amy oppong''}\\
& \orexample{``Write a music-hall song called "The Parlor Upstairs" ...''}\\

\rowcolor{gray!20}
\textsc{generate joke} & \orexample{``You are a Radio DJ at 99.7 Now Radio. Write me a funny radio break about presenting the next radio program...''}\\
\rowcolor{gray!20}
& \orexample{``tell me some jokes with a naval or military theme which can be used in a farwell speech''}\\

\textsc{question about writing} & \orexample{``could a person with potential with lots of friends ... be an effective Villain backstory? How can it be written well, and how can it be written poorly?''}\\
& \orexample{``The following excerpt is from an alternate history novel. Can you critique the system of government described in the excerpt?...''}\\
\bottomrule
\end{tabular}
}
\caption{Example original requests from \wildchatwr for each writing intent type.} 
\label{tab-fgtype-examples}
\end{table*}

\end{document}